\documentclass[final,3p,times,twocolumn,authoryear]{elsarticle}
\pdfoutput=1

\PassOptionsToPackage{draft}{hyperref}
\usepackage{graphicx}
\usepackage[cmex10]{amsmath}
\usepackage{array}
\usepackage{amsfonts}
\usepackage{gensymb}
\usepackage{color}
\usepackage{lineno}
\usepackage{subfigure}
\usepackage{eqnarray}
\usepackage{multirow}
\usepackage{todonotes}

\newcolumntype{L}[1]{>{\raggedright\let\newline\\\arraybackslash\hspace{0pt}}m{#1}}
\newcolumntype{C}[1]{>{\centering\let\newline\\\arraybackslash\hspace{0pt}}m{#1}}
\newcolumntype{R}[1]{>{\raggedleft\let\newline\\\arraybackslash\hspace{0pt}}m{#1}}

\biboptions{comma,round}

\newcommand{\sourcecodeurl}{\url{https://github.com/ignaciorlando/red-lesion-detection}}

\journal{Computer Methods and Programs in Biomedicine}

\begin{document}

\begin{frontmatter}

\title{An Ensemble Deep Learning Based Approach for Red Lesion Detection in Fundus Images}

\author[pladema,conicet]{Jos\'{e}~Ignacio~Orlando\corref{mycorrespondingauthor}}
\cortext[mycorrespondingauthor]{Corresponding author}
\ead{jiorlando@conicet.gov.ar}

\author[wiv-isp,famhp]{Elena~Prokofyeva}

\author[pladema,cic]{Mariana~del~Fresno}

\author[esat]{Matthew~B.~Blaschko}

\address[pladema]{Pladema Institute, UNCPBA, Gral. Pinto 399, Tandil, Argentina}
\address[conicet]{Consejo Nacional de Investigaciones Cient\'ificas y T\'ecnicas, CONICET, Argentina}
\address[cic]{Comisi\'on de Investigaciones Cient\'ificas de la Provincia de Buenos Aires, CIC-PBA, Buenos Aires, Argentina}
\address[wiv-isp]{Scientific Institute of Public Health (WIV-ISP), Brussels, Belgium}
\address[famhp]{Federal Agency for Medicines and Health Products (FAMHP), Brussels, Belgium}
\address[esat]{ESAT-PSI, KU Leuven, Kasteelpark Arenberg 10, B-3001 Leuven, Belgium}

\begin{abstract}

\noindent \textbf{Background and objectives:} Diabetic retinopathy (DR) is one of the leading causes of preventable blindness in the world. Its earliest sign are red lesions, a general term that groups both microaneurysms (MAs) and hemorrhages (HEs). In daily clinical practice, these lesions are manually detected by physicians using fundus photographs. However, this task is tedious and time consuming, and requires an intensive effort due to the small size of the lesions and their lack of contrast. Computer-assisted diagnosis of DR based on red lesion detection is being actively explored due to its improvement effects both in clinicians consistency and accuracy. Moreover, it provides comprehensive feedback that is easy to assess by the physicians. Several methods for detecting red lesions have been proposed in the literature, most of them based on characterizing lesion candidates using hand crafted features, and classifying them into true or false positive detections. Deep learning based approaches, by contrast, are scarce in this domain due to the high expense of annotating the lesions manually. 

\noindent \textbf{Methods:} In this paper we propose a novel method for red lesion detection based on combining both deep learned and domain knowledge. Features learned by a convolutional neural network (CNN) are augmented by incorporating hand crafted features. Such ensemble vector of descriptors is used afterwards to identify true lesion candidates using a Random Forest classifier.

\noindent  \textbf{Results:} We empirically observed that combining both sources of information significantly improve results with respect to using each approach separately. Furthermore, our method reported the highest performance on a per-lesion basis on DIARETDB1 and e-ophtha, and for screening and need for referral on MESSIDOR compared to a second human expert. 

\noindent \textbf{Conclusions:} Results highlight the fact that integrating manually engineered approaches with deep learned features is relevant to improve results when the networks are trained from lesion-level annotated data. An open source implementation of our system is publicly available at \sourcecodeurl.
\end{abstract}



\end{frontmatter}


\section{Introduction}
\label{sec:introduction}

One of the most common consequences of vascular damage due to diabetes mellitus is Diabetic Retinopathy (DR), which is one of the leading causes of preventable blindness in the world \citep{ProkofyevaZrenner2012}. As the prevalence of diabetes worldwide is expected to increase from 2.8\% to 4.4\% from 2000 to 2030, and about 5\% of people with Type-2 diabetes have DR, it is expected that the number of patients suffering from this disease will significantly increase in the next years \citep{abramoff2015mass}.

One of the earliest signs of DR are microaneurysms (MAs), which are balloon-shaped deformations on the vessel walls, induced by the permeability of the vasculature due to hyperglycemia \citep{mookiah2013computer}. While DR progresses, the number of MAs increases, and some of them can break and produce leakages of blood on the retinal layers, namely hemorrhages (HEs)\footnote{In some clinical literature, the acronym HE stands Hard Exudates. However, we use it here to refer to hemorrhages, in line with the biomedical computing literature \citep{seoud2015red}.}. The most commonly used term to refer to both MAs and small HEs is ``red lesions'' \citep{niemeijer2005automatic,decenciere2013teleophta,seoud2015red}. The accumulation of blood or lipids induce swelling, which can result in retinal damage when it reaches the macula and, potentially, blindness \citep{abramoff2010retinal}. 

\begin{figure}
\label{fig:subtle-red-lesions}
\includegraphics[width=\columnwidth]{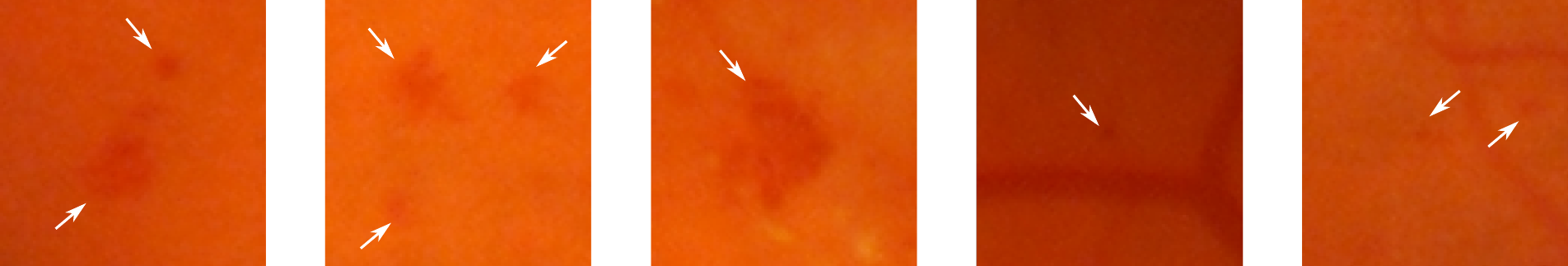}
\caption{Examples of red lesions observed in fundus photographs from DIARETDB1 \citep{kauppi2007diaretdb1}.}
\end{figure}

In its early stages, DR might be clinically asymptomatic \citep{abramoff2010retinal}. As a consequence, this condition is typically identified when it is more advanced and treatments are significantly less effective \citep{mookiah2013computer}. A recent study has shown that 44\% of hospitalized patients with diabetes remain undiagnosed \citep{kovarik2016prevalence}. To prevent this, people suffering from diabetes are usually recommended to be regularly examined through fundus images to verify the non-existence of red lesions \citep{abramoff2010retinal}.
Although fundus photographs are currently the most economical non-invasive imaging technique for this purpose, manual diagnosis requires an intensive effort to screen the images \citep{mookiah2013computer}. Red lesions appear as small red dots that might be subtle and too small to be detected at first glance (Figure~\ref{fig:subtle-red-lesions}). Large HEs, on the contrary, are more evident and less difficult to visualize.

Automated methods for computer-aided diagnosis are known to significantly reduce the time, cost, and effort of DR screening: their high throughput ensures the more efficient analysis of large populations \citep{sanchez2011evaluation}. They also reduce the intra-expert variability, which is commonly high due to the small size and the irregular shape of the lesions \citep{abramoff2015mass}. These systems are usually aided by an automated module for red lesion detection. In general, the problem of red lesion detection is tackled using a two-stage approach, consisting first of detecting a set of potential candidates, and then refining this set with a classifier trained using hand crafted features \citep{niemeijer2005automatic,walter2007automatic,niemeijer2010retinopathy,seoud2015red}. 

Convolutional Neural Networks (CNNs) have recently emerged as a powerful framework to solve a large variety of computer vision and medical image analysis problems \citep{krizhevsky2012imagenet,zheng20153d,venkataramani2016understanding}. Such methods are able to learn features automatically from a sufficiently large training set, without requiring the manual design of the filters. CNNs are known to outperform other manually engineered approaches on a large variety of applications \citep{razavian2014cnn}. Their discrimination ability is usually affected by the amount of available training data: deeper architectures are known to be able to learn more discriminative features, although at the cost of requiring larger data sets to prevent overfitting and ensure a proper generalization error \citep{goodfellow2016deep}. Image level annotations of large scale data sets can be obtained in a relatively economical way~\citep{trucco2013validating}. However, labeling images at a lesion level is costly, tedious and time consuming, as it requires the intervention of experienced experts who must zoom within different areas of the images to identify every single pathological structure, as accurately as possible. This fact significantly influences the performance of deep learning based approaches for red lesion detection, which must be trained using lesion level annotated data.


In this study we propose to take advantage of both deep learned and manual engineered features for red lesion detection in fundus photographs. In particular, we propose to learn a first set of discriminative features using a light CNN architecture, and then augment their original characterization ability by incorporating hand crafted descriptors. These ensemble vectors of features are used to train a Random Forest classifier that is applied at test time to discriminate between true and false lesion candidates. We experimentally observed that the deep learned features are complementary to the manually engineered, and are aided by the incorporation of domain knowledge. 


\subsection{Related works}
\label{sec:related-works}

Deep learning methods for DR screening have significantly attracted the attention of the research community after the release of the Kaggle competition database,\footnote{\url{https://kaggle.com/c/diabetic-retinopathy-detection}} which provides a large amount of fundus photographs with image-level annotations. Recently, \cite{gulshan2016development} have presented a CNN that achieved impressive performance for detecting patients with moderate/worse and severe/worse DR. The output of such a method is a quantitative indicator of the risk of the patient's being at a moderate or advanced stage of DR. Red lesion detection methods, by contrast, are intended to identify earlier stages of the disease, providing probability maps that indicate the location of its clinical signs. This feature allows physicians to visually assess the correctness of the results, while helping them to achieve a more reliable and accurate early diagnosis.

Red lesion detection in fundus photographs have been extensively explored in the literature, although most of the existing approaches are based on detecting MAs or HEs separately, and not both structures simultaneously \citep{niemeijer2010retinopathy,van2016fast,seoud2015red}. Moreover, current existing approaches are based exclusively on hand crafted features. This is likely due to the fact that deep learning based methods have to be trained from large data sets with lesion level annotations. This setting has direct implications on why deep learning based models have been ignored for tackling the problem of red lesion detection. One exception is the method for HEs detection by~\cite{van2016fast}. 
This approach is focused on detecting HEs at different scales, which are in general more evident than MAs. By contrast, our method is used for detecting both MAs and small HEs simultaneously, which are more difficult to be visually assessed by physicians.

As mentioned above, in this study we present an ensemble approach that improves the features learned by a CNN by incorporating domain knowledge. Only few efforts have been made in the literature to analyze the viability of such an approach. 
\cite{annunziata2016accelerating}, for instance, propose to initialize a convolutional sparse coding approach with manually designed filters to accelerate its learning process and improve their original discriminative power. That approach is applied for detecting curvilinear structures such as neurons or retinal vessels, which are easier to manually trace.  
\cite{venkataramani2016understanding} have observed that state of the art descriptors significantly improve the performance of transferred CNN features when applied to kidney detection in ultrasound images. The main difference with respect to our approach is that our CNN is trained from scratch from a domain specific data set, while the approach of \cite{venkataramani2016understanding} is based on fine-tunning a CNN trained from natural images.

From our literature review, we identified two main methods resembling our approach, although with different applications and based on different CNN architectures. \cite{zheng20153d} introduce a method for identifying landmarks in 3D CT scans using the output of a dedicated CNN in combination with Haar features to boost the quality of the results. Its deep learning based component is divided into two stages: a first stage, based on a light architecture with only one hidden layer, is used to recover a large set of landmark candidates; the second stage, made up of three hidden layers and trained using sparsity priors, is used to recover a large vector of neural network features, which is combined with Haar features to train a probabilistic boosting-tree classifier. In order to save as much data as possible for training the CNN and the lesion classifier, we avoided performing candidate detection in a supervised way. Instead, a combination of morphological operations and image processing techniques is used to retrieve potential lesions, without using training data. This allows us to train a slightly deeper architecture in the subsequent stage, only dedicated to classifying the lesion candidates, which is able to capture discriminative features from the training patches.

The method by \cite{wang2014mitosis} for mitosis detection on histopathology images is also similar to ours. It uses candidate detection as well, and a RF classifier trained using hand crafted features is applied to assign a probability of being a true mitosis candidate. In parallel, a CNN with two convolutional layers and one fully connected layer is trained from patches around the candidates to retrieve an additional probability. The final decision is performed via consensus of the predictions of the two classifiers by weighting both probabilities using two manually tuned parameters. We took the alternative approach of using both feature vectors simultaneously to train the RF classifier, as it can take advantage of the interaction between both the deep learned and the hand crafted features.

\subsection{Contributions}

In this paper we propose to learn discriminative models for red lesion detection by combining both deep learned and hand crafted features. First, an unsupervised, candidate detection approach based on morphological operations is applied to retrieve a set of potential lesions. Next, a CNN is trained from a set of patches around the candidate lesions to learn a first feature vector. These descriptors are augmented with a set of hand crafted features to improve their ability to distinguish the true positive lesions. A Random Forest (RF) classifier is trained using this hybrid feature vector, and is then applied for refining the set of candidates, discriminating between true lesions and false positives. We empirically observed that combining both sources of information improved performance not only when evaluating our method on a per-lesion basis but also when analyzing its potential for DR screening or need for referral detection on an image-level basis. Our results on benchmark data sets such as e-ophtha \citep{decenciere2013teleophta}, DIARETDB1 \citep{kauppi2007diaretdb1} and MESSIDOR \citep{decenciere2014feedback} show that our strategy outperforms other state of the art methods that are not only based on red lesion detection but also in detecting other pathological structures such as exudates or neovascularizations. An extensive analysis of the complementarity of the deep learned features with respect to the hand crafted ones is also provided, with the purpose of assessing their contribution in the discrimination process.

\section{Methods}
\label{sec:methods}

\begin{figure*}[t]
	\centering
	\includegraphics[width=\textwidth]{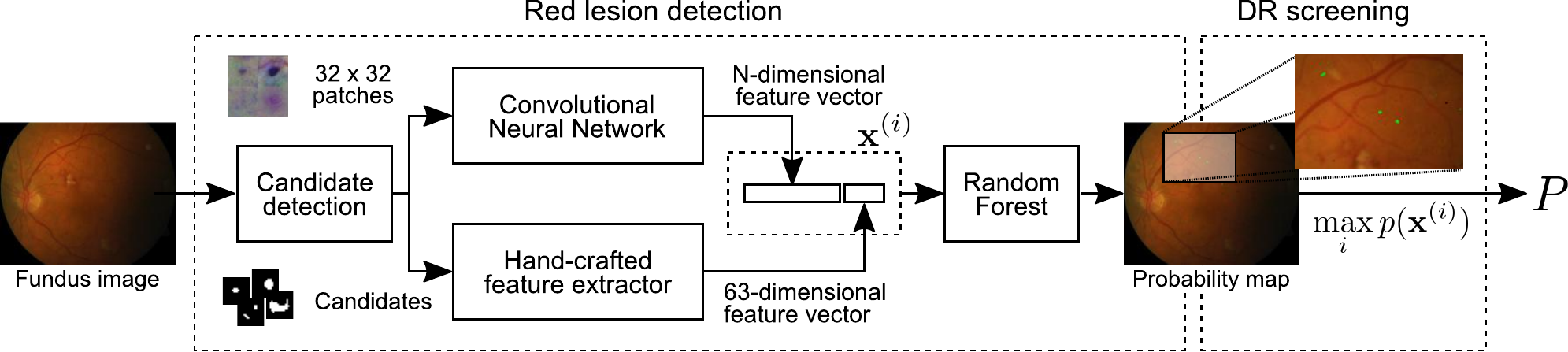} 
	\caption{Overview of our method for red lesion detection.}
	\label{fig:overview}
\end{figure*}

A schematic representation of our method is depicted in Figure~\ref{fig:overview}. Lesion candidates retrieved with morphological operations (Section~\ref{subsec:candidate-detection}) are filtered using a set of hybrid descriptors. Regular patches centered on each candidate connected component are collected to build a training set that is used to train a CNN (Section~\ref{subsec:cnn}). A 63-dimensional vector of hand crafted features (Section~\ref{subsec:hand-crafted}) is also computed per each of the candidates. A Random Forest (RF) classifier (Section~\ref{subsec:classification}) is afterwards trained on the resulting combination of features, and used to classify the new candidates. Since the presence of red lesions is the first indicator of DR, the maximum over lesion likelihoods is used to assign a DR probability, as done by~\cite{seoud2015red} and \cite{antal2012ensemble}. 

\subsection{Candidate detection}
\label{subsec:candidate-detection}

\begin{figure*}[t]
	\centering
	\includegraphics[width=\textwidth]{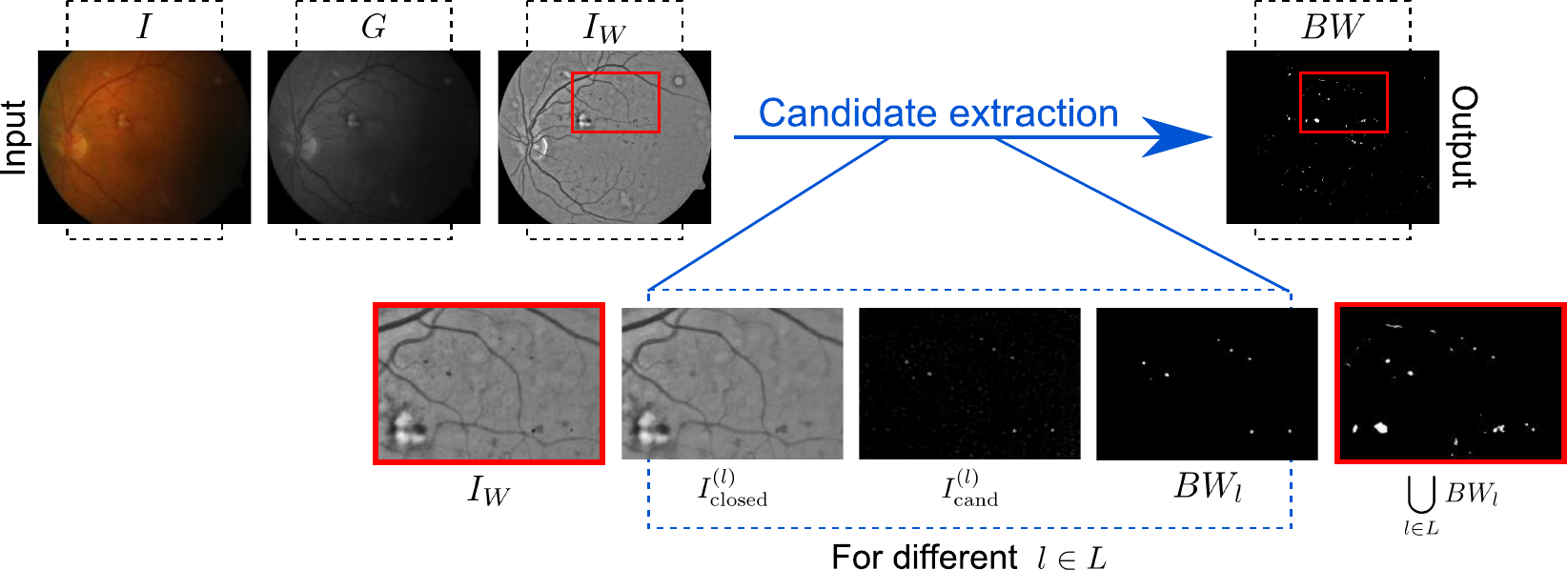} 
	\caption{Red lesion candidate detection.  See Section~\ref{subsec:candidate-detection} for a detailed description of the process.}
	\label{fig:candidates}
\end{figure*}

\begin{figure}[t]
  \centering
  \subfigure[Original RGB image]{\includegraphics[width=0.3\columnwidth]{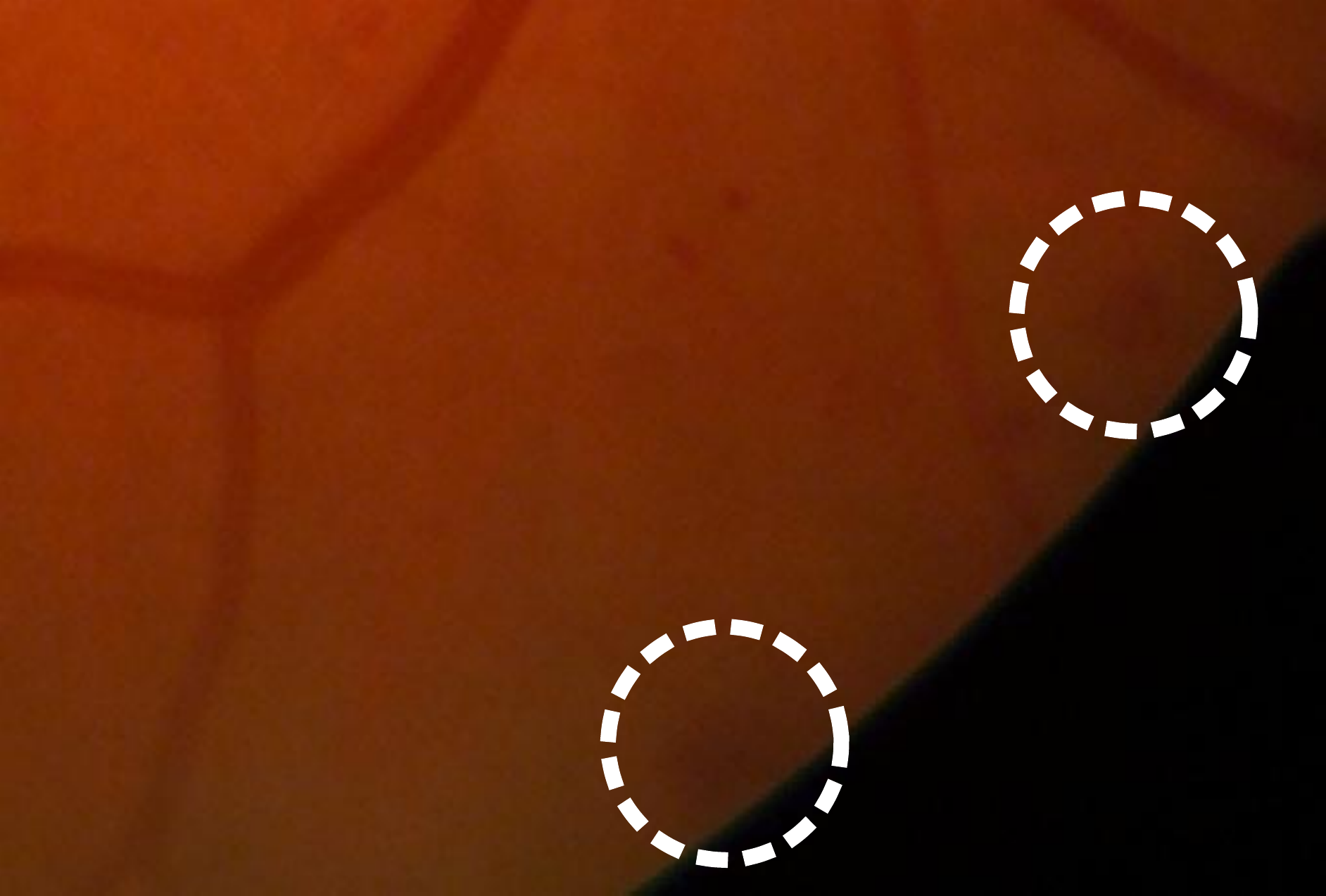}\label{fig:i-w-rgb}}
  \subfigure[$I_W$ from $G$]{\includegraphics[width=0.3\columnwidth]{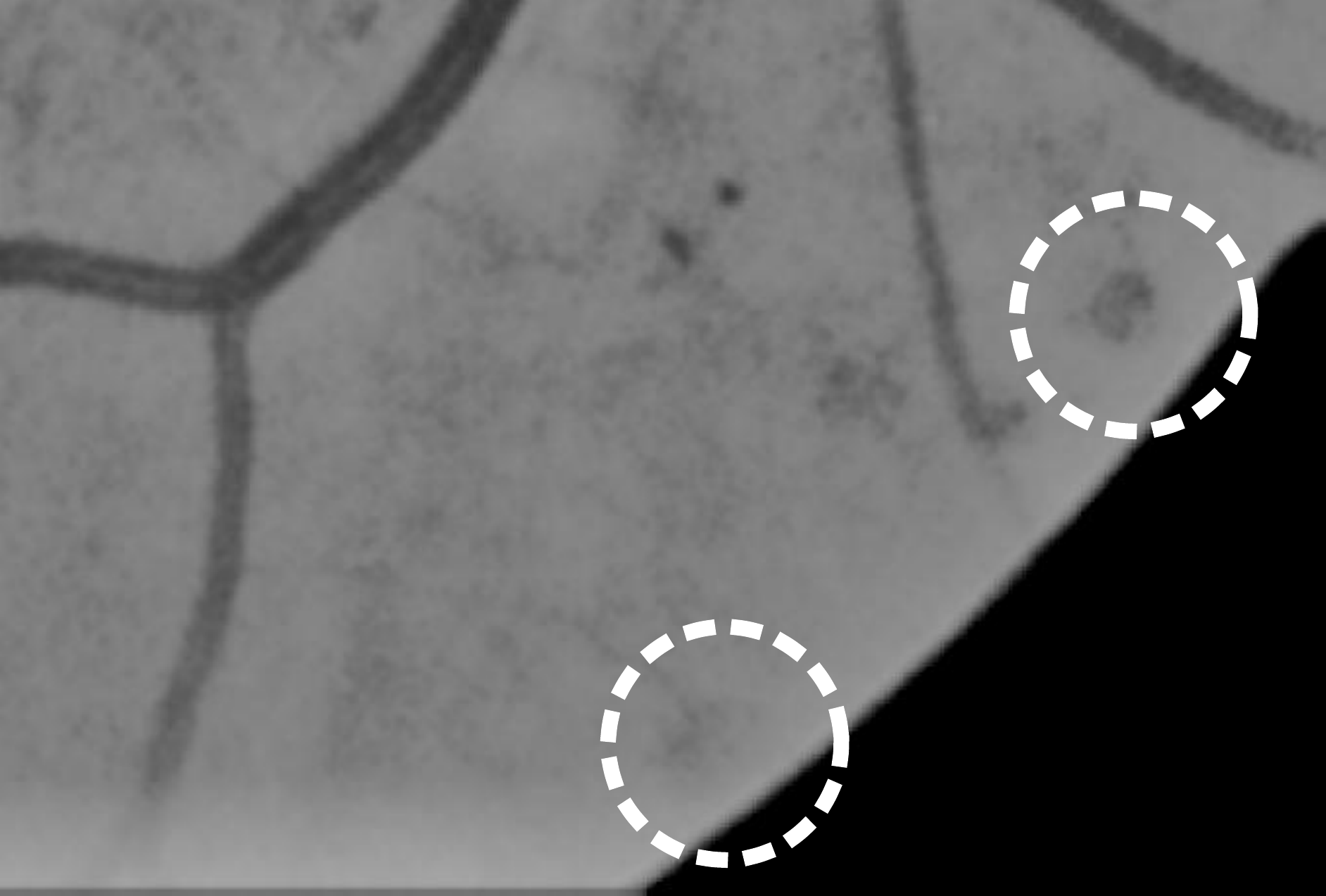}\label{fig:i-w-without}}
  \subfigure[$I_W$ from expanded $G$]{\includegraphics[width=0.3\columnwidth]{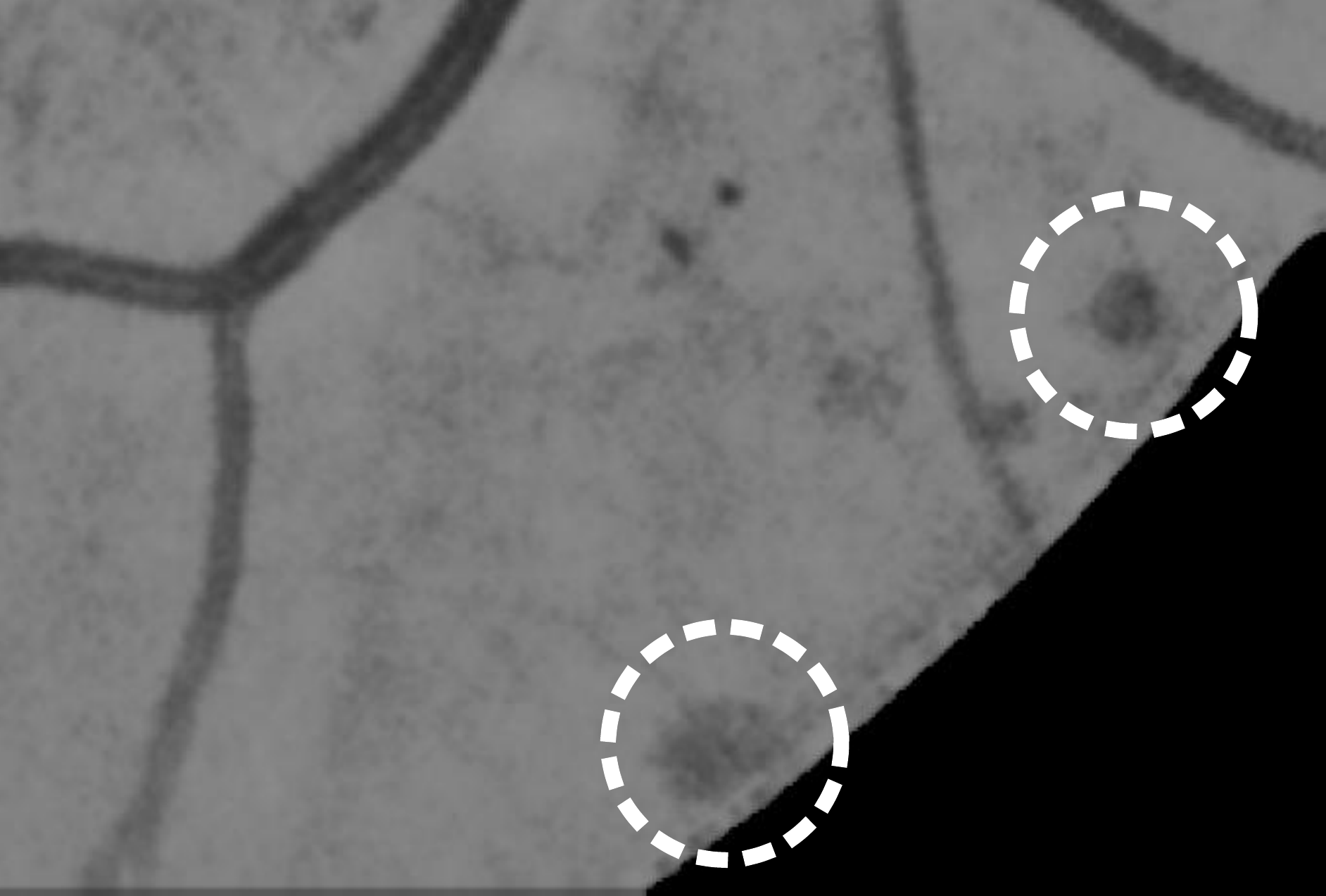}\label{fig:i-w-with}}
  \caption{Effect of the FOV expansion on the lesion candidates located closely to the border of the FOV.}
  \label{fig:artifacts}
\end{figure}

Our strategy for candidate detection is illustrated in Figure~\ref{fig:candidates}. First, the green band $G$ from the original color image $I$ is taken, since it is the one that allows a better visual discrimination of the red lesions. To avoid artifacts in the borders of the FOV that might hide potential lesions (Figure~\ref{fig:i-w-without}), a wider aperture of $\frac{3}{30} \mathcal{X}$ pixels is simulated~\citep{soares2006retinal} from $G$, where $\mathcal{X}$ corresponds to the width in pixels of the field of view (FOV). Since our purpose is to develop a system sufficiently general to be applied at different image resolutions, all the relevant parameters are expressed in terms of $\mathcal{X}$.

As uneven background illumination might hide potential lesions occurring within the darkest areas of the images, a $r$-polynomial transformation is applied on pixel intensities:
\begin{equation}
\label{eq:walterTransformation}
I_{\text{W}}(i,j)= \left\{ \begin{array}{lr}
             \frac{\frac{1}{2}(u_\text{max} - u_\text{min})}{{(\mu_W(i,j) - \min{(G)})}^r}, 				 & G(i,j) \leq \mu_W(i,j) \\
             \\ \frac{-\frac{1}{2}(u_\text{max} - u_\text{min})}{{(\mu_W(i,j) - \max{(G)})}^r}, 		 & G(i,j) > \mu_W(i,j)
             \end{array}
   \right.
\end{equation}
with $r=2$, $u_\text{min}=0$ and $u_\text{max}=1$, respectively, and where $\mu_W$ is the local average intensity on square neighborhoods of length $W$, computed for each $(i,j)$ pixel~\citep{walter2007automatic}. We observed that using $W=25$ performed sufficiently well for enhancing images with $536$ pixels of horizontal resolution such as those in the DRIVE data set \citep{niemeijer2004comparative}, so this parameter is automatically adjusted using $W = \frac{25}{536} \mathcal{X}$. Figure~\ref{fig:artifacts} illustrates how the expansion of the FOV border and the subsequent intensity transformation improve the contrast of subtle lesions located in the border of the FOV.

A Gaussian filter with $\sigma=5$
is applied to $I_{\text{W}}$ to reduce noise, resulting in a new image $I'_{\text{W}}$. Afterwards, different morphological closings are performed on $I'_{\text{W}}$ using linear structuring elements of length $l \in L$ at angles $\theta$ spanning from $0$ to $180^\circ$ with increments of $15^\circ$. The set of relevant scales $L$ is a fixed parameter that is also automatically adjusted in terms of $\mathcal{X}$, as explained in Section~\ref{subsec:model-selection}. By taking the minimum response over all the considered angles, an image $I^{(l)}_{\text{closed}}$ is obtained in which responses to lesions with sizes smaller than $l$ were reduced, and all the remaining structures are still preserved~\citep{walter2007automatic}. A score map is then obtained by:
\begin{equation}
\label{eq:icand}
I^{(l)}_{\text{cand}} = I^{(l)}_{\text{closed}} - I_{\text{W}}.
\end{equation}
Afterwards, a thresholding operation is applied on $I^{(l)}_{\text{cand}}$, where the threshold is automatically determined in such a way that a maximum of $K=120$ candidates are retrieved from the score map. In order to achieve this goal, thresholds $t_s$ from $\min{(I^{(l)}_{\text{cand}})}$ to $\max{(I^{(l)}_{\text{cand}})}$ with increments of 0.002 are explored until the number of connected components in the resulting binary maps is less than or equal to $K$. To support the cases in which no lesions are detected or when is not possible to detect less than $K$ candidates, a lower bound $t_l$ and an upper bound $t_u$ are experimentally set such that:
\begin{equation}
\label{eq:thresholding}
t_K = \left\{ \begin{array}{lr}
						t_l, & \forall t_s : \text{CC}(I^{(l)}_{\text{cand}}>t_s) < K \\
            			t_k, & \text{CC}(I^{(l)}_{\text{cand}}>t_s) \leq K \\
						t_u,	 & \forall t_s : \text{CC}(I^{(l)}_{\text{cand}}>t_s) > K \\
             \end{array}
   \right.
\end{equation}
where $\text{CC}$ is a function than counts the number of connected components in the thresholded score map. Once $t_K$ is fixed, a binary map of candidates is obtained by thresholding $BW_l = I^{(l)}_{\text{cand}} > t_K$~\citep{walter2007automatic}. This operation is repeated for different values of $l \in L$ to capture potential lesions at different scales, so the binary map of candidates $BW$ is obtained as $BW = \bigcup_{l \in L} BW_l$. Finally, as $BW$ might include small candidates which usually are not associated to any pathological region but with noise, all connected structures in $BW$ with less than $px$ pixels are discarded. The automated model selection procedure used to set the values of $K$ and $px$ and the scales in $L$ is described in Section~\ref{subsec:model-selection}.

Figure~\ref{fig:cnn-training-set} presents a random sample of the potential lesions retrieved by the method on a randomly selected image from DIARETDB1 training set. It is possible to see that most of false positive samples correspond to vascular branching or crossing points, vessel segments and beadings, scars due to laser photocoagulation or black spots of dirt in the capture device, as reported by \cite{seoud2015red}. This setting underlines the importance of refine the candidates to remove false positives.

\subsection{CNN-based features}
\label{subsec:cnn}

We train a dedicated CNN to characterize each red lesion candidate. For this purpose, each color band of the original image $I$ is equalized first as proposed by~\cite{van2016fast}:
\begin{equation}
	I_\text{ce}(i,j;\sigma) = \alpha \cdot I(i,j) + \tau \cdot \text{Gaussian}(i,j;\sigma) * I(i,j) + \gamma
\end{equation}
where $*$ is a convolution, the Gaussian filter has a standard deviation $\sigma = \frac{\mathcal{X}}{30}$, and $\alpha=4$, $\tau=-4$ and $\gamma=128$ were set following \cite{van2016fast}. We empirically observed that this preprocessing operation not only dramatically diminishes the number of epochs needed for training but also improves the discrimination ability of the CNN. As explained in Section~\ref{subsec:candidate-detection}, a wider FOV is also simulated for each color band 
to prevent any undesired effect in the FOV border.  

\begin{figure}[t!]
  \centering
  \subfigure[Non-lesions (false positive candidates)]{\includegraphics[width=0.7\columnwidth]{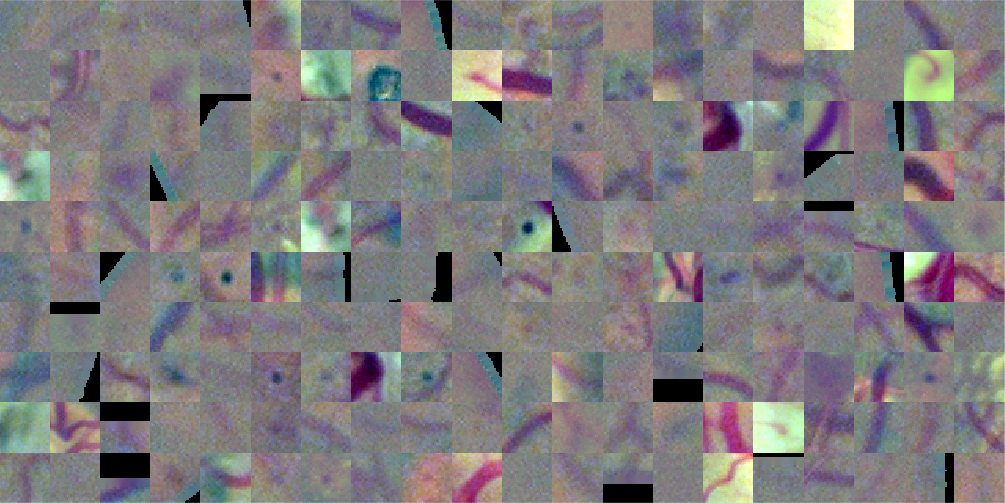}\label{fig:false-positives}}
  
    \subfigure[Lesions (true positive candidates)]{\includegraphics[width=0.7\columnwidth]{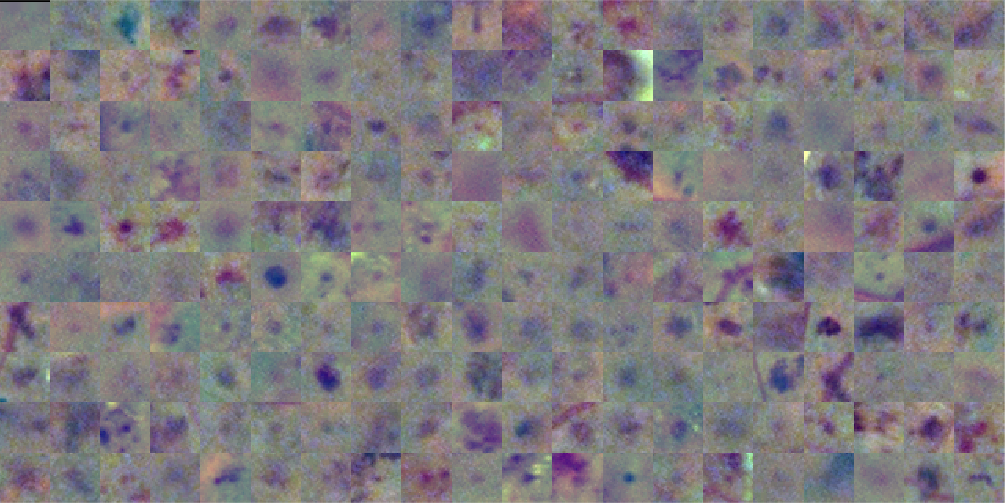}\label{fig:true-positives}}
	\caption{CNN training set. Random sample of 200 patches for (a) non lesions and (b) true lesions.  See Section~\ref{subsec:cnn} for details of the construction of the training set.}
	\label{fig:cnn-training-set}
\end{figure}

A training set $S = \{(X^{(i)}, y^{(i)})\}, i = 1, ..., n$ is built for training the CNN, where each sample $X^{(i)}$ is a square patch around the center of each red lesion candidate, as extracted from $I_\text{ce}$ (Figure~\ref{fig:cnn-training-set}). The patch size is taken as double the length of the major axis of the candidate, or $32 \times 32$ pixels if the major axis of the candidate is smaller than 32 pixels. This setting let us to recover not only the candidate itself but also its surrounding area, which allows the CNN to capture both candidates' internal features and information about its shape, borders and context. Patches larger than $32 \times 32$ pixels are downsized to this resolution to ensure a uniform input size for the CNN. As windows are square by definition, this transformation is isotropic and does not affect the appearance of the lesion candidate. 
Samples are centered by subtracting the training set mean image. Alternative scaling methods such as ZCA whitening and contrast normalization were also analyzed, although no improvements were observed on the validation set when applying them.
The label $y^{(i)} \in \{0, 1\}$ associated to the candidate is assigned according to the ground truth labeling on the data set: if the candidate overlaps with a true labeled region, then the window is assumed to be a true red lesion ($y = 1$); if it does not, then it is assumed to be a false positive ($y = 0$). The CNN is trained from scratch on an $8\text{x}$ augmented version of this training set, which is obtained by rotating each patch by $90^\circ, 180^\circ$ or $270^\circ$, and then flipping the resulting windows horizontally or not. Thus, for each input patch, 8 new patches are generated.

The CNN architecture is depicted in Table~\ref{table:architecture}. It comprises 4 convolutional layers and 1 fully connected layer with 128 units. This layer is used to retrieve the $N = 128$-dimensional vector of deep learned features. The CNN was designed by using the original LeNet architecture as the initial baseline, and introducing changes by evaluating their contribution on reducing the empirical error on a held-out validation set, which was randomly sampled from the training set. Using deeper architectures such as VGG-S or Inception-V3 was avoided as the increase in the number of parameters would have required a larger training set to reduce overfitting.

Depending on the number of images in the training set with an advanced level of DR and the size of the lesions we focus on detecting, the classification problem is imbalanced to a greater or lesser degree. Hence, the proportion of true positive lesions might be significantly smaller that the number of false positive ones. If this imbalance grows dramatically, it was previously observed that the typical cross-entropy loss is affected, and, as a consequence, fewer true positives are retrieved~\citep{maninis2016deep}. Thus, we used a class balanced cross-entropy loss, given by:
\begin{align}\label{eq:ClassBalancedCrossEntropyLoss}
\mathcal{L}_\beta(\mathbf{W}) =& -\beta \sum_{i \in Y_+} \log P(y^{(i)} | X^{(i)}; \mathbf{W}) \\ &- (1 - \beta) \sum_{i \in Y_-} \log P(y^{(i)} | X^{(i)}; \mathbf{W})) \nonumber 
\end{align}
where $\mathbf{W}$ are the weight parameters of the network; $P$ is the probability obtained by applying a sigmoid function to the activation of the fully connected layer; $Y_+$ and $Y_-$ are the subsets of true and false positive samples, respectively; and $\beta = |Y_-| / (|Y_+| + |Y_-|)$ is the ratio of negative vs. positive samples in $S$.

The CNN's weights were randomly initialized from a Gaussian distribution with a lower standard deviation for the first layer (0.01) than for the remaining ones (0.05), to prevent vanishing gradients. Standard dropout after all the convolutional layers was analyzed as an alternative to the reported architecture, although it was observed that it did not improve results on the validation set. Moreover, using such an approach increased the training time significantly. We noticed instead that using dropout after the first convolutional layer with a high keep probability $p=0.99$ slightly improved results. We also used weight decay of $5 \times 10^{-4}$ for regularization, to penalize large $\mathbf{W}$ values during backpropagation. Batch normalization was also evaluated but no improvements were observed when applying it. The cost function was optimized using stochastic gradient descent, with a batch size of 100 samples taken from the training set, which is randomly shuffled at the beginning of each epoch. The learning rate was initially fixed in $\eta=0.05$, and it was divided by a factor of 2 every time that the relative improvement in current $\mathcal{L}_\beta(\mathbf{W})$ value was less than $1\%$ of the average loss over the last 10 epochs. The optimization process was stopped when this relative difference  was smaller than $10^{-4}$, or a maximum number of 200 epochs was achieved. The CNN was implemented in MATLAB R2015b, using Matconvnet~\citep{vedaldi2015matconvnet}. To improve performance during training, a NVIDIA Titan X GPU card was used, achieving convergence in 20-30 minutes.

\begin{table}[t]
\centering
\caption{CNN architecture. Convolutional layers (conv) indicate width, height and depth of each learned filter. Pooling layers (pool) include the dimension of the pooling operation and the stride. Dropout is only applied after the first convolutional layer with a low dropout probability.}
\label{table:architecture}
\resizebox{\columnwidth}{!}{
\begin{tabular}{c|c|c|c}
  \hline
  \textbf{Block} & \textbf{Layers} & \textbf{Filter size} & \textbf{Output size} \\
  \hline
  \multirow{3}{*}{1} & conv    & $5 \times 5 \times 3$ & 32 \\
  				     & maxpool & $3 \times 3$ - $\text{stride}=2$ & \\	  
                     & dropout & $p = 0.01$	& \\
  \hline
  \multirow{2}{*}{2} & conv    & $5 \times 5 \times 32$ & 32 \\
  				     & avgpool & $3 \times 3$ - $\text{stride}=2$ & \\	  
  \hline
  \multirow{2}{*}{3} & conv    & $5 \times 5 \times 32$ & 64 \\
  				     & avgpool & $3 \times 3$ - $\text{stride}=2$ & \\	           \hline
  4 				 & conv    & $4 \times 4 \times 64$ & $N$ \\
  \hline
  5 				 & fully connected  & $N$ & $N$ \\
  \hline
  6 				 & $\mathcal{L}_\beta (\mathbf{W})$ &  $N$ & 2 \\             
  \hline
\end{tabular}
}
\end{table}


\subsection{Hand-crafted feature extraction}
\label{subsec:hand-crafted}

As a complementary source of information with respect to the CNN features, a 63 dimensional feature vector of hand-crafted features (HCF) is extracted per each lesion candidate and incorporated to our feature vector. Some of these descriptors were extensively explored in the literature~\citep{niemeijer2005automatic,niemeijer2010retinopathy,seoud2015red}, while other are introduced here to improve the existing ones. In general, they can be divided into two categories: intensity based and shape based features (Table~\ref{table:hand-crafted-description}). 

Intensity features exploit the visual properties of the candidate areas, and are extracted from different versions of the color image $I$, obtained by applying different preprocessing strategies. In particular, we extracted descriptors used in the state of the art \citep{niemeijer2010retinopathy, seoud2015red} but from the following derived images: 
\begin{itemize}
\item Original red, green and blue color bands ($R$, $G$ and $B$, respectively).
\item Green band $G$ after illumination correction ($I_W$, obtained as in Section~\ref{subsec:candidate-detection}).
\item Color bands and $I_W$ after CLAHE contrast enhancement ($R_c$, $G_c$, $B_c$, ${I_W}_c$).
\item Color bands after color equalization ($R_\text{ce}$, $G_\text{ce}$, $B_\text{ce}$).
\item $I_\text{SC}$, which is the difference between the green band $G$ and an estimated background $I_\text{BG}$, obtained using a median filter with squared windows of length $\frac{25}{536} \mathcal{X}$.
\item $I_\text{match}$. This image is obtained by initially computing $I_\text{lesion}$, which is a vessel free version of $I_\text{SC}$, obtained by inpainting the vasculature as in~\citep{orlando2017convolutional}. The difference between each pixel $(i,j)$ in $I_\text{lesion}$ and its $11 \times 11$ neighborhood is assigned to $I_\text{match}(i,j)$. 
\item $I_\text{cand} = \max_l I^{(l)}_{\text{cand}}$, which is the maximum response to the candidate score map described in Section~\ref{subsec:candidate-detection}, taken from $I_W$, but restricting the size of the structuring elements to the lengths $l \in \{5, 7, ..., 15\}$. 
\end{itemize}

Shape based features have the ability to characterize the structure of the candidates. Red lesions are expected to be relatively circular, with small area and perimeter, and approximately equal minor and major axis. Such statistics, including compactness, eccentricity and aspect ratio, are also included as part of the domain knowledge feature vector. 

\begin{figure}[t!]
  \centering
  \subfigure[]{\includegraphics[width=0.3\columnwidth]{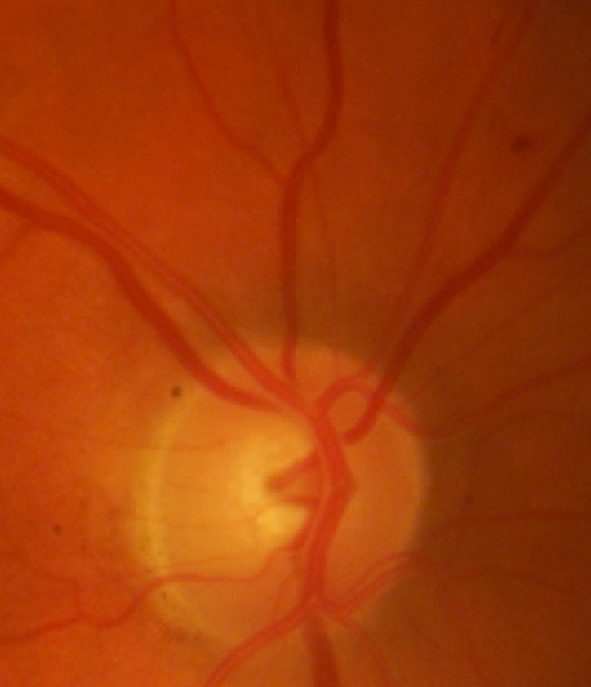}\label{fig:rgb-vessel}}
  \subfigure[]{\includegraphics[width=0.3\columnwidth]{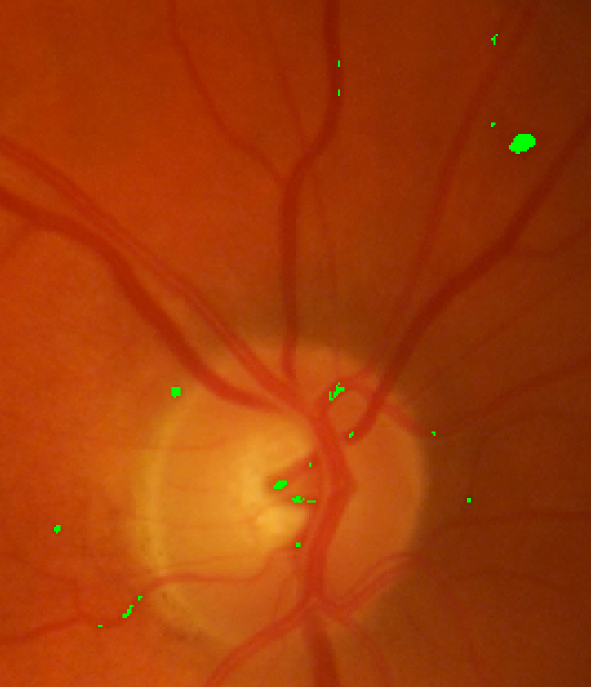}\label{fig:rgb-vessel-cand}}
  \subfigure[]{\includegraphics[width=0.3\columnwidth]{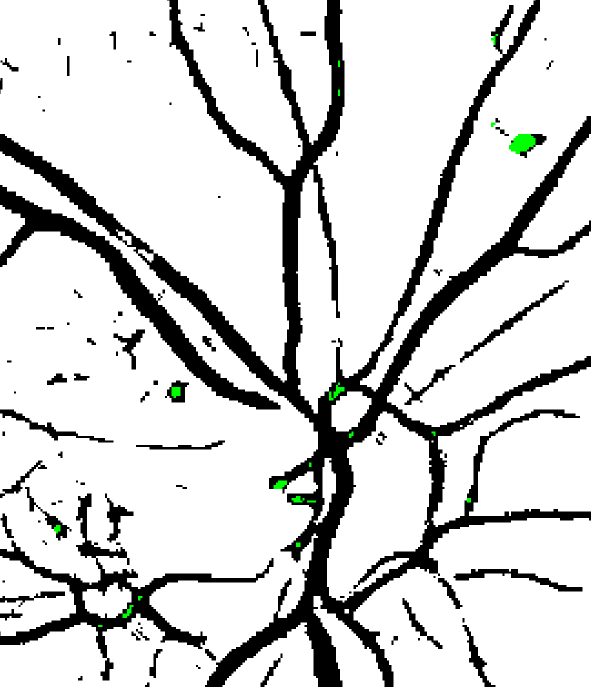}\label{fig:vessel-segm-cand}}
  
  \subfigure[]{\includegraphics[width=0.3\columnwidth]{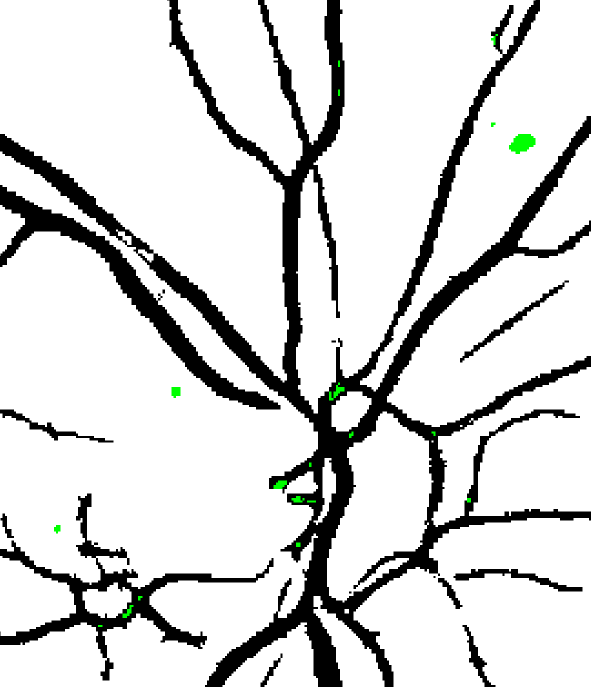}\label{fig:vessel-segm2-cand}}
  \subfigure[]{\includegraphics[width=0.3\columnwidth]{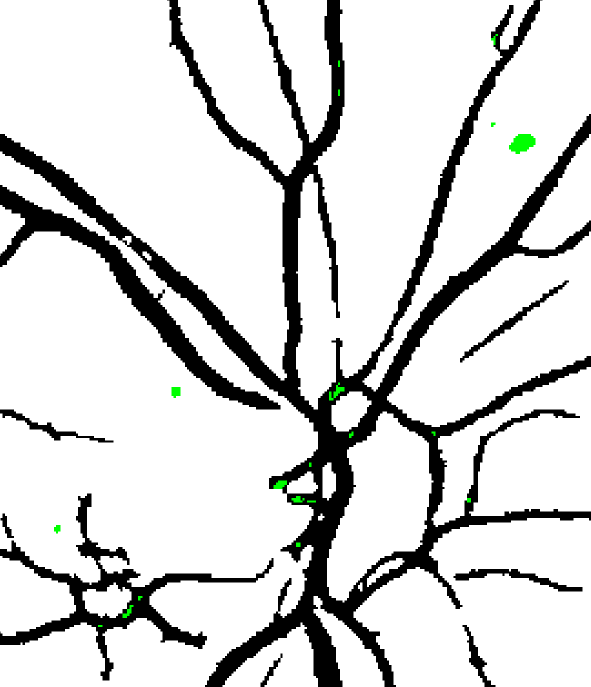}\label{fig:vessel-segm3-cand}}
  \caption{Feature based on vessel segmentation. (a) $I$. (b) $I$ with candidates superimposed. (c) Vessel segmentation. (d) Vessel segmentation after removing spurious elements. (e) Vessel segmentation after morphological closing.}
  \label{fig:segmentation-based-feature}
\end{figure}

We also analyzed the viability of using the segmentation of the retinal vasculature as a potential source of information. As seen in Figure~\ref{fig:false-positives}, most of the false positive detections are located in vessel crossings or beadings. Thus, we compute an initial vessel segmentation using the method reported in~\citep{orlando2014learning,orlando2016discriminatively}, and postprocessing the output by removing every spurious connected component with less than $\frac{100}{536} \mathcal{X}$ pixels \cite{orlando2017convolutional}. A morphological closing with a disk of radius 2 is afterwards applied to fill any gap due to the central reflex in arteries. Then, we measure the ratio of pixels in the candidate region that overlap with the segmentation, divided by the number of pixels in the candidate. Figure~\ref{fig:segmentation-based-feature} illustrates the process of computing this feature. It can be seen that most of the false positive lesions located at the optic disc overlap with the resulting segmentation mask, and can be removed by this descriptor.

\begin{table*}
\centering
\caption{Summary of the hand crafted features used to complement our CNN. Top: intensity based features. Bottom: shape based features.}
\resizebox{\textwidth}{!}{
\begin{tabular}{ m{13cm} | C{4cm}  }
  \hline                       
  \textbf{Intensity based features (dimensionality)} & \textbf{Extracted from} \\
  \hline \hline
  Average intensity value in the candidate region. (13) & $R$, $G$, $B$, $I_W$, $R_c$, $G_c$, $B_c$, ${I_W}_c$, $R_{\text{ce}}$, $G_{\text{ce}}$, $B_{\text{ce}}$, $I_{\text{SC}}$,$I_\text{top-hat}$\\
  \hline
  Sum of intensities in the candidate region. (12)	& $R$, $G$, $B$, $I_W$, $R_c$, $G_c$, $B_c$, ${I_W}_c$, $R_{\text{ce}}$, $G_{\text{ce}}$, $B_{\text{ce}}$, $I_{\text{SC}}$	  \\
  \hline
  Standard deviation of intensities in the candidate region. (12)	&	$R$, $G$, $B$, $I_W$, $R_c$, $G_c$, $B_c$, ${I_W}_c$, $R_{\text{ce}}$, $G_{\text{ce}}$, $B_{\text{ce}}$, $I_{\text{SC}}$\\
  \hline
  Contrast: Difference between mean intensity in the candidate region and mean intensity of the dilated region (12)	&	$R$, $G$, $B$, $I_W$, $R_c$, $G_c$, $B_c$, ${I_W}_c$, $R_{\text{ce}}$, $G_{\text{ce}}$, $B_{\text{ce}}$, $I_{\text{SC}}$ \\
  \hline
  Normalized total intensity: Difference between total and mean intensities of the candidate area in $I_{\text{BG}}$, divided by the candidate's standard deviation in $I_{\text{BG}}.$ (3) & $G$, $I_{\text{SC}}$, $I_W$	\\
  \hline
  Normalized mean intensity: Difference between mean intensity in $I_W$ and mean intensity of the candidate area in $I_{\text{BG}}$, divided by the standard deviation of the candidate in $I_{\text{BG}}$. (1) & $I_W$	\\
  \hline
  Minimum intensity in the candidate area. (1) & $I_{\text{match}}$ \\
  \hline
  \hline                   
  \textbf{Shape based features (dimensionality)} & \textbf{Extracted from} \\
  \hline  
  \hline  
   Area: Number of pixels of the candidate. (1) & $BW$  \\
  \hline
  Perimeter: Number of pixels on the border of the candidate. (1) & $BW$  \\ 
  \hline
  Aspect ratio: Ratio between the major and minor axis lengths. (1) & $BW$  \\
    \hline
  Circularity $ = 4 \pi \text{Area} / \text{Perimeter}^2 $. (1) & $BW$  \\
  \hline
  Compactness $ = \sqrt{(\sum_{j=1}^n d_j - \tilde{d}) / n}$, where $d_j$ is the distance from the centroid of the object to its $j$th boundary pixel and $\tilde{d}$ is the mean of all the distances from the centroid to all the edge pixels. $n$ is the number of edge pixels. (1) & $BW$  \\
  \hline
  Major axis of the ellipse that has the same normalized second central moments as the candidate region. (1) & $BW$ \\
  \hline
  Minor axis of the ellipse that has the same normalized second central moments as the candidate region. (1) & $BW$ \\
  \hline
  Eccentricity of the ellipse that has the same second-moments as the candidate region. The eccentricity is the ratio of the distance between the foci of the ellipse and its major axis length. (1) & $BW$ \\
  \hline
  Ratio of the pixels on the candidate region that are also included in the binary segmentation of the retinal vasculature, obtained as in~\citep{orlando2017convolutional}. (1) & Vessel segmentation \\
  \hline
\end{tabular}
}
\label{table:hand-crafted-description}
\end{table*}

\subsection{Candidate classification with Random Forest}
\label{subsec:classification}

A Random Forest (RF) is an ensemble classifier that is widely used in the literature due to its capability to perform both classification and feature selection simultaneously~\citep{Breiman2001,lovercio2017detection}. It is also robust against overfitting, which is relevant when having small training sets, and is suitable to deal with noisy, high dimensional imbalanced data. We trained this classifier for the purpose of refining our set of candidates using our hybrid feature vector. In all our experiments, we standardized the features to zero mean an unit variance.


A RF is a combination of $T$ decision trees. These trees are learned from $T$ examples that are randomly sampled with replacement from our training set $S$. Each node in a tree corresponds to a split made using the best of a randomly selected subset of $m = \sqrt{d}$ features, with $d$ being the dimensionality of the feature vector. The quality of the split is given by the decrease in the Gini index that the split produces~\citep{Breiman2001}. Given a feature vector $\mathbf{x}^{(j)}$, the RF evaluates the conditional probability $p_i(c|\mathbf{x}^{(j)})$, where $c \in \{-1, 1\}$ is the class--with -1 corresponding to a non lesion and 1 to a true lesion-- and $i$ is the index of the tree in the forest. The final probability is then computed by repeating this process for every tree $0 < i \leq T$, and averaging the responses of each of them:
\begin{equation}
p(c|\mathbf{x}^{(j)}) = \frac{1}{T} \sum_i^T p_i(c|\mathbf{x}^{(j)})
\end{equation}

In order to determine the probability $P$ of the image $I$ corresponding to a DR patient or not, we followed the same procedure used by~\cite{seoud2015red}:
\begin{equation}
P(I) = \max_j p(c=1|\mathbf{x}^{(j)}),
\label{eq:max-prob}
\end{equation}
which means that for a given image $I$ with $m$ lesion candidates, the probability of being DR will be associated with the maximum certainty of the classifier of having observed a true positive lesion ($c=1$).

\section{Experimental setup}
\label{sec:experiments}


\subsection{Materials}
\label{subsec:materials}

  We conducted experiments using three publicly available data sets: DIARETDB1\footnote{\url{http://www.it.lut.fi/project/imageret/diaretdb1/}}
  ~\citep{kauppi2007diaretdb1}, e-ophtha\footnote{\url{http://www.adcis.net/en/Download-Third-Party/E-Ophtha.html}}
  ~\citep{decenciere2013teleophta}, and MESSIDOR\footnote{\url{http://messidor.crihan.fr}.}
  ~\citep{decenciere2014feedback}

\begin{table}[t!]
\centering
\caption{Distribution of DR grades in the MESSIDOR data set, and diagnostic criterion. MA $=$ microaneurysms, HE $=$ hemorrhages and NV$=$ neovascularizations.}
\label{table:messidor}
\footnotesize\rm
\resizebox{\columnwidth}{!}{
\begin{tabular}{c|c|c}
\hline               
\textbf{Grade} 	& \textbf{Criteria} & \textbf{Num. images} \\
\hline
\hline
R0 & ($N_\text{MA}=0$) AND ($N_\text{HE}=0$) &  546   \\
R1 & ($0 < N_\text{MA} \leq 5$) AND ($N_\text{HE}=0$)	& 153	\\
R2 & ($5 < N_\text{MA} < 15$) AND ($0 < N_\text{HE} < 5$) AND ($N_\text{NV}=0$) & 247	\\
R3 & ($N_\text{MA} \geq 15$) OR ($N_\text{HE} \geq 5$) OR ($N_\text{NV}>0$) 	& 254	\\
\hline
\end{tabular}
}
\end{table}

DIARETDB1 and e-ophtha were used to perform a per-lesion evaluation as they provide lesion level annotations. MESSIDOR provides image level annotations indicating the DR grade, assigned using the criterion detailed in Table~\ref{table:messidor}. Thus, this set was used to quantify the performance of our method as a DR screening tool, on a per-image basis. We also used e-ophtha for this purpose, by generating image-level annotations based on the number of red lesions in the ground truth segmentation. Thus, any image with at least one red lesion was labeled as DR. The ROC\footnote{\url{http://webeye.ophth.uiowa.edu/ROC/}}
~\citep{niemeijer2010retinopathy} training set, which comprises 50 fundus photographs taken at different resolutions, was used to augment DIARETDB1 training set for small red lesion detection on e-ophtha. Further details about the experimental setup are provided in Table~\ref{table:experiments}.

DIARETDB1 consists of 89 color fundus images taken under varying imaging settings~\citep{kauppi2007diaretdb1}. 84 images contain signs of mild or pre-proliferative DR, and the remaining 5 are considered normal. The entire set is divided into a training set and a test set of 28 and 61 images, respectively. Four different experts have delineated the regions where MA and HE can be found, and a consensus map is provided per each type of lesion. The standard practice is to evaluate MA or HE detection methods at a conservative $\geq 75\%$  agreement~\citep{kauppi2007diaretdb1}. For red lesion detection, however, \cite{seoud2015red} propose to use as ground truth the union of the consensus maps for both MAs and HEs at a $> 25\%$ level of agreement. We followed this latter approach to evaluate our red lesion detection strategy.

e-ophtha~\citep{decenciere2013teleophta} is a database generated from a telemedical network for DR screening, and it includes manual annotations of MAs and small HEs. It comprises 148 images with small red lesions, and 233 with no visible sign of DR. In order to obtain per-image labels indicating the presence or absense of DR, images with any red lesion were labeled as DR.

Finally, MESSIDOR~\citep{decenciere2014feedback} comprises 1200 color fundus images acquired by 3 ophthalmic institutions in France. Images were originally captured at different resolutions, and graded into four different DR stages, being R0 the healthy category and R3 the most severe. Two different classification problems are usually derived from MESSIDOR grades: DR screening, which corresponds to distinguishing R0 from the remaining R1, R2 and R3 grades~\citep{antal2012ensemble,seoud2015red}; and detecting the need for referral, which corresponds to R0 and R1 vs. R2 and R3 grades~\citep{sanchez2011evaluation,pires2013assessing}. We evaluated our method on a per image basis following these two approaches. 

Since these data sets do not include FOV masks, which are necessary for processing the images, we automatically generate them by thresholding the luminosity plane of the CIELab version of the RGB images at 0.15 (for DIARETDB1, e-ophtha and MESSIDOR) and 0.26 (for ROC)~\citep{orlando2017convolutional}. If the resulting binary mask is such that the entire image is estimated as a foreground, an alternative approach is applied where the RGB bands are summed up and the resulting image is thresholded at an empirically tuned value of 150. To smooth borders and reduce noise, all masks are postprocessed with a median filter using square windows of side 5, and only its largest connected component is preserved.  In principle, these masks would be available directly from the fundus camera, and the process of replicating this information directly from the images is a necessary but not central task to the present paper. The FOV masks for all the data sets used in this paper are released in the project webpage (Section~\ref{sec:conclusions}).


\begin{table*}
\centering
\caption{Experimental setup. 
$\beta$ is the value for the balanced cross-entropy loss (Equation~\eqref{eq:ClassBalancedCrossEntropyLoss}).}
\label{table:experiments}
\resizebox{0.9\textwidth}{!}{
\begin{tabular}{ C{1cm} || C{2.2cm} | C{2.5cm} | C{2.5cm} | C{1.5cm} | C{1.5cm} | C{1cm} ||  C{2.5cm} | C{2.3cm}}
  \hline                       
  \textbf{Exp. ID} & \textbf{Detection} & \textbf{Training set} & \textbf{GT labels} & \textbf{True lesions} & \textbf{Non lesions} & \textbf{$\beta$} & \textbf{Per lesion evaluation} & \textbf{Per image evaluation} \\
  \hline
  \hline
  1 & Red lesions with multiple sizes & DIARETDB1 training set (28 images) & $\text{MA}~>~25\% \cup \text{HE}~>25~\%$ & 1059 (27\%) & 2905 (73\%) & $\beta = 0.5$ & DIARETDB1 test set & MESSIDOR \\
  \hline
  2 & Small red lesions & DIARETDB1 \& ROC training sets (78 images) & $\text{MA}>75\%$ from DIARETDB1 \& ROC MA labels & 407 (4\%) & 10282 (96\%) & $\beta \sim 0.96$ & e-ophtha & e-ophtha \\
  \hline
\end{tabular}}
\end{table*}

\subsection{Model selection}
\label{subsec:model-selection}

Candidate detection relies on three significant parameters: $L$, which is the set of scales used to retrieve potential candidates; $K$, the number of candidates retrieved for a given scale; and $\text{px}$, the minimum area in pixels that a candidate must have. In our experiments, these values were experimentally adjusted using the DIARETDB1 training set, resulting in $L = \{3, 6, 9, \dots , 60\}$, $K=120$ and $px=5$. The maximum scale from $L$ was adapted on the remaining data sets using a scaling factor of $\frac{\mathcal{X}}{1425}$, where 1425 is the average width of the images in DIARETDB1. This allows to recover a set of candidates with a size proportional to the resolution of each image. 

The parameters of the CNN (in particular, dropout probability $1-p$ and the size of the fully connected layer $N$) were designed according to the performance on a held out validation set, randomly sampled from each training set. The parameters that maximized the area under the precision/recall curve ($N=128$ and $p=0.99$) were always used for evaluation on the test set. The number of trees $T \in \{ 100, 120, ..., 200 \}$ for the RF was fixed to the value that minimized the out-of-bag error on the training set on each experiment~\citep{Breiman2001}. The maximum number of possible trees was fixed to a relatively low value (200) to reduce the computational cost during training and prediction. Nevertheless, experiments adding up to 2000 trees to the model did not show any improvements in reducing the out-of-bag error.

\subsection{Evaluation metrics}
\label{subsec:evaluation-metrics}

Free-response ROC (FROC) curves were used to evaluate the performance of our red lesion detection method on a per lesion basis. These plots, which are extensively used in the literature to estimate the overall performance on this task, represent the per lesion sensitivity against the average number of false positive detections per image (FPI) obtained on the data set for different thresholds applied to the candidate probabilities. Thus, FROC curves provide a graphical representation of how the model is able to deal with the detection of true lesions in all the images of the data set. We also computed the Competition Metric (CPM) as proposed in the Retinopathy Online Challenge~\citep{niemeijer2010retinopathy}, which is the average per lesion sensitivity at the reference FPI values $\in \{ 1/8, 1/4, 1/2, 1, 2, 4, 8\}$. The protocol used by~\cite{seoud2015red} was followed when evaluating in DIARETDB1, as indicated in Section~\ref{subsec:materials}. 

When evaluating on a per image basis, we used standard ROC curves, where both the sensitivity ($Se = \frac{TP}{FN + TP}$) and $1 - \text{specificity}$ ($Sp = \frac{TN}{FP + TN}$) are depicted within the same plot for different DR probability values, obtained as indicated in Equation~\ref{eq:max-prob}. Additionally, we studied the $Se$ at $Sp=50\%$, which is a standard comparison metric for screening systems~\citep{sanchez2011evaluation}.

\section{Results}
\label{sec:results}

\subsection{Per lesion evaluation}
\label{subsec:per-lesion-evaluation}

Two different experiments were conducted for per lesion evaluation, as detailed in Table~\ref{table:experiments}. FROC curves are used for comparison, and Wilcoxon signed rank tests were performed to estimate the statistical significance of the differences in the per lesion sensitivity values. These tests were conducted using 100 sensitivity values retrieved for logarithmically spaced FPI values in the interval $[\frac{1}{8}, ..., 8]$, which corresponds to a more dense version of the reference FPI values used for computing the CPM~\citep{niemeijer2010retinopathy}.

\begin{figure}[t!]
  \centering
  \includegraphics[width=0.9\columnwidth]{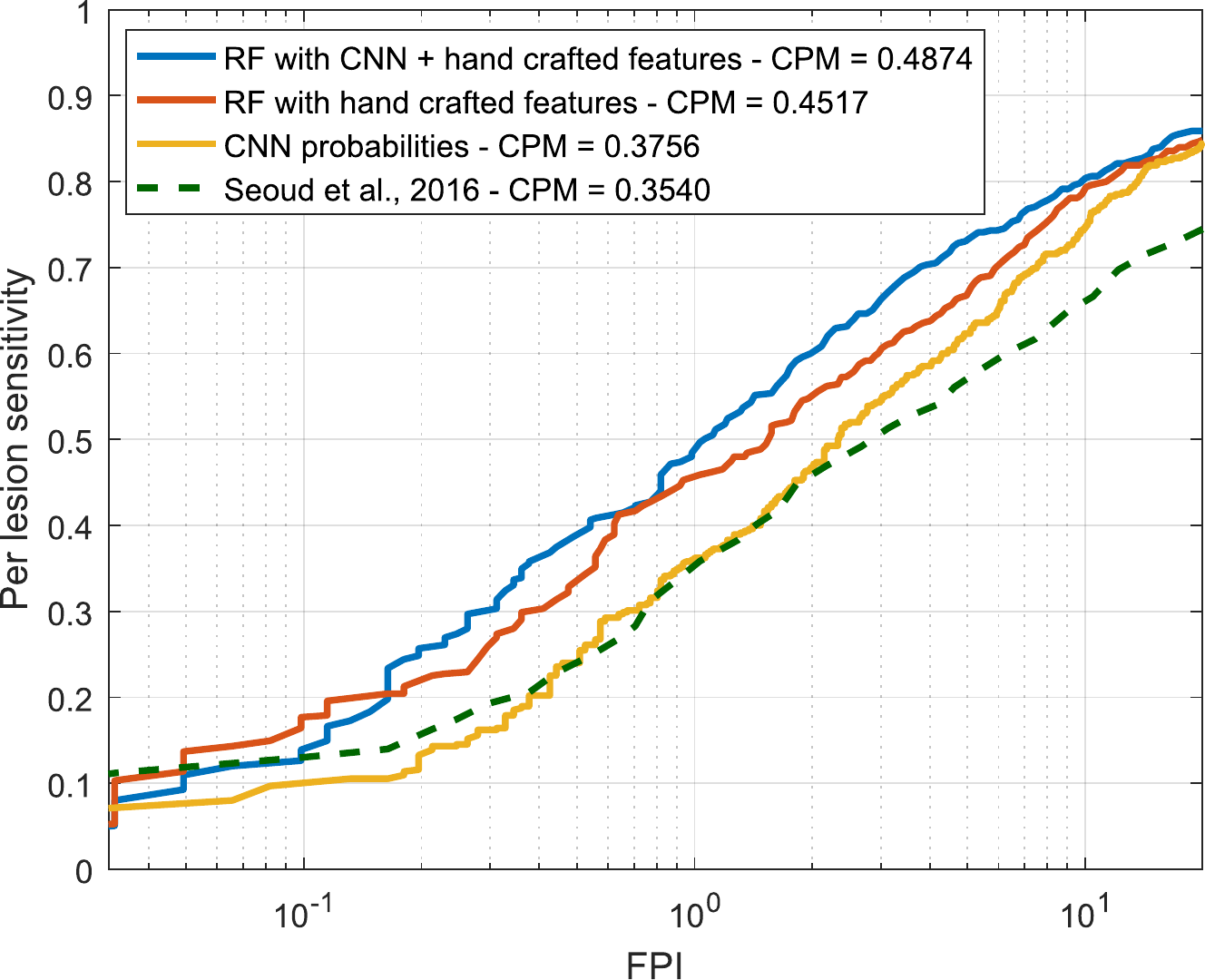} 
  \caption{Per lesion evaluation in Experiment 1. FROC curve and CPM values obtained on the DIARETDB1 test set.}
	\label{fig:diaretdb1}
\end{figure}

Experiment 1 evaluates the model ability to deal with both MAs and HEs simultaneously at multiple scales, following the same protocol as \cite{seoud2015red} (Figure~\ref{fig:diaretdb1}). Results obtained by \cite{seoud2015red} were provided by the authors and obtained using the same training and test configuration, and are included for comparison purposes. Hypothesis tests show a statistically significant improvement in the per lesion sensitivity values when using the combined approach compared to using each representation separately ($p < 2 \times 10^{-18}$ and $p < 4 \times 10^{-17}$ for the CNN probabilities and the hand crafted features, respectively). Moreover, the hybrid method reported better results compared to \citeauthor{seoud2015red} ($p < 2 \times 10^{-18}$).

\begin{figure}[t!]
  \centering
  \subfigure[Microaneurysms]{\includegraphics[width=0.7\columnwidth]{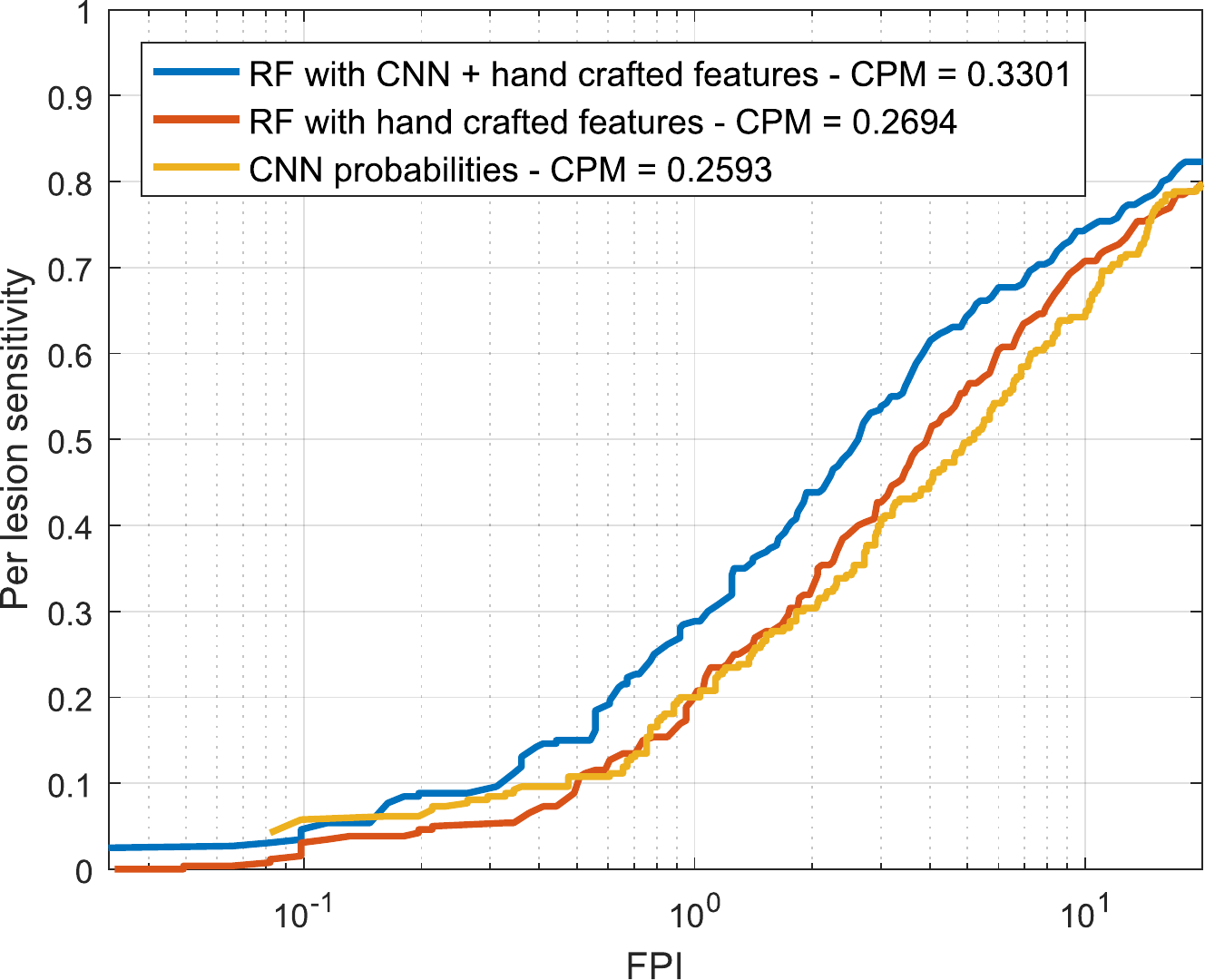}\label{fig:lesion-microaneurysms}}
  
  \subfigure[Hemorrhages]{\includegraphics[width=0.7\columnwidth]{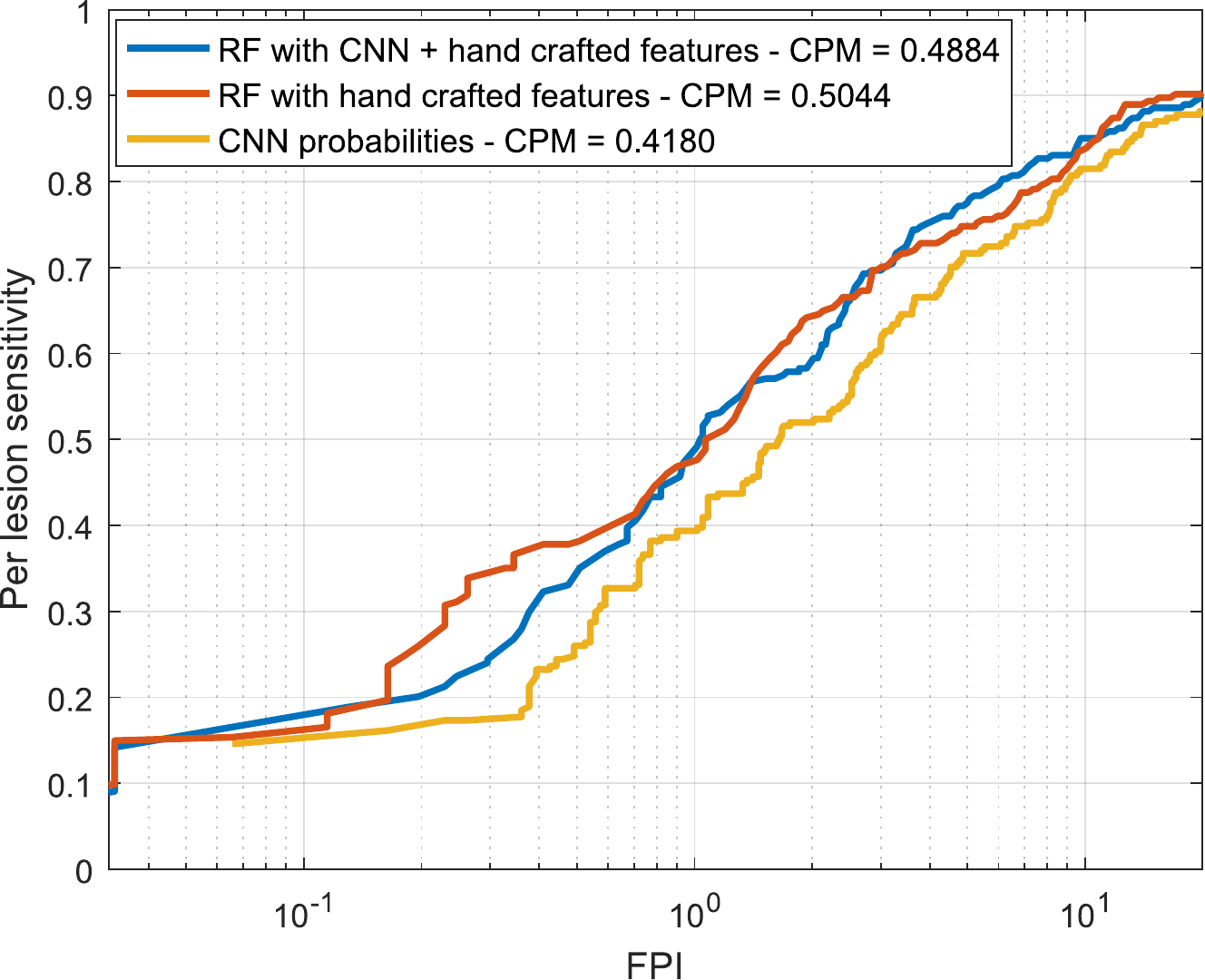}\label{fig:lesion-hemorrhages}}
  \caption{Per lesion evaluation for each lesion type in the DIARETDB1 test set: (a) Microaneurysms, (b) Hemorrhages.}
  \label{fig:per-lesion-type}
\end{figure}

As DIARETDB1 includes labels for both MAs and HEs, it is possible to quantitatively assess the accuracy of the method to detect each type of lesion. Figure~\ref{fig:per-lesion-type} illustrates the FROC curves and the CPM values obtained by the models learned in Experiment 1, when analyzing MAs and HEs separately. For MA detection, the combined approach achieves higher per lesion sensitivity values than using each approach separately ($p < 2 \times 10^{-18}$ and $p < 3 \times 10^{-17}$ for the hand crafted features and the CNN, respectively), with a noticeable improvement at the clinically relevant FPI=1 value (0.2885 versus 0.202 and 0.2 for combined, CNN, and hand crafted, respectively). Moreover, the differences between the manually tuned approach and the CNN probabilities are not statistically significant. 
When evaluating the ability of the system to detect HEs on the DIARETDB1 test set, it is possible to see that the per lesion sensitivities are higher than those reported for MA detection. Furthermore, the hand crafted features are able to achieve better per lesion sensitivity values than the combined approach ($p < 5 \times 10^{-5}$) for this specific task. At the clinically relevant FPI value of 1, however, the combined approach reports a slightly higher per lesion sensitivity compared to the manually engineered descriptors (0.4907 versus 0.4724).

\begin{figure}[t!]
  \centering
  \includegraphics[width=0.9\columnwidth]{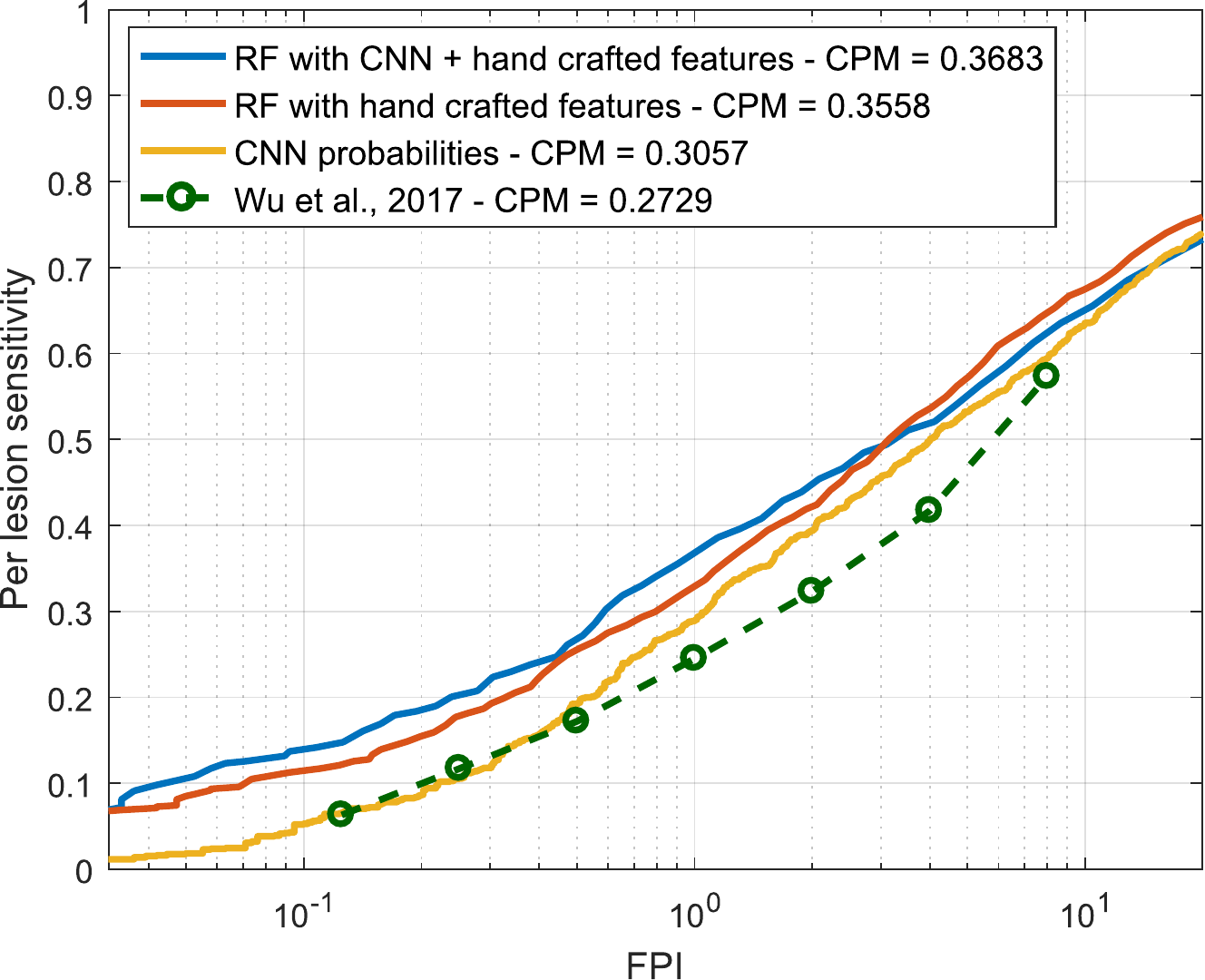}
\caption{Per lesion evaluation in Experiment 2. FROC curve and CPM values obtained on e-ophtha.}
	\label{fig:e-ophtha}
\end{figure}

Experiment 2 was carried out on e-ophtha to estimate the ability of our method to segment MAs and smaller HEs simultaneously. In this case, a combination of both the DIARETDB1 (MA labels with a level of agreement $\geq 75\%$) and ROC training sets was used for learning, as we observed that few MAs (only 182 for the entire DIARETDB1 set) are retrieved at $\geq 75\%$ agreement. To the best of our knowledge, the only method evaluated on e-ophtha is by~\cite{wu2017automatic}, although their analysis is performed on a subsample of 74 images with lesions instead of the full data set. By contrast, we used a more challenging evaluation comprising the entire e-ophtha set, including also the 233 images with no visible sign of DR. Figure~\ref{fig:e-ophtha} presents the FROC curves obtained using each approach. As in the previous experiment, the Wilcoxon signed rank tests showed a statistical significant improvement in the per lesion sensitivity values using the hybrid vector of both deep learned features and domain knowledge with respect to the CNN probabilities and the hand crafted features ($p < 2 \times 10^{-18}$ and $p < 2 \times 10^{-9}$, respectively).

Table~\ref{table:cpm-values-and-se} summarizes the CPM values obtained for each experiment and each feature combination, and also using each of the two recently published state-of-the-art methods. Per lesion sensitivities at FPI$=1$, which is considered a clinically relevant number of false positives \citep{niemeijer2010retinopathy} are also provided.

\begin{table}[t!]
\centering
\caption{CPM values and per lesion sensitivities at FPI$=1$ for Experiments~1 (red lesions with multiple sizes) and~2 (small red lesions) (Table~\ref{table:experiments}).}
\label{table:cpm-values-and-se}
\resizebox{\columnwidth}{!}{\begin{tabular}{ c || c | c || c | c }
  \hline
  \multirow{2}{*}{\textbf{Method}} & \multicolumn{2}{c||}{\textbf{Experiment 1}} & \multicolumn{2}{c}{\textbf{Experiment 2}} \\

  & CPM & $Se$ & CPM & $Se$  \\
  \hline
  \hline
  \cite{seoud2015red} & 0.3540 & 0.3462 & - 	   & - \\
  \hline
  \cite{wu2017automatic} & - 	 & - 	  & 0.2729 & 0.2450\\
  \hline
  CNN probabilities    	& 0.3756 & 0.3621 & 0.3057 & 0.2894\\
  \hline
  RF with HCF 			& 0.4517 & 0.4601 & 0.3558 & 0.3291 \\
  \hline
  \textbf{RF with CNN + HCF} 	& \textbf{0.4874} & \textbf{0.4883} & \textbf{0.3683} & \textbf{0.3680} \\
  \hline
\end{tabular}}
\end{table}

\begin{figure}[t!]
 	\centering
    \subfigure[DIARETDB1 test image]{\includegraphics[width=0.47\columnwidth]{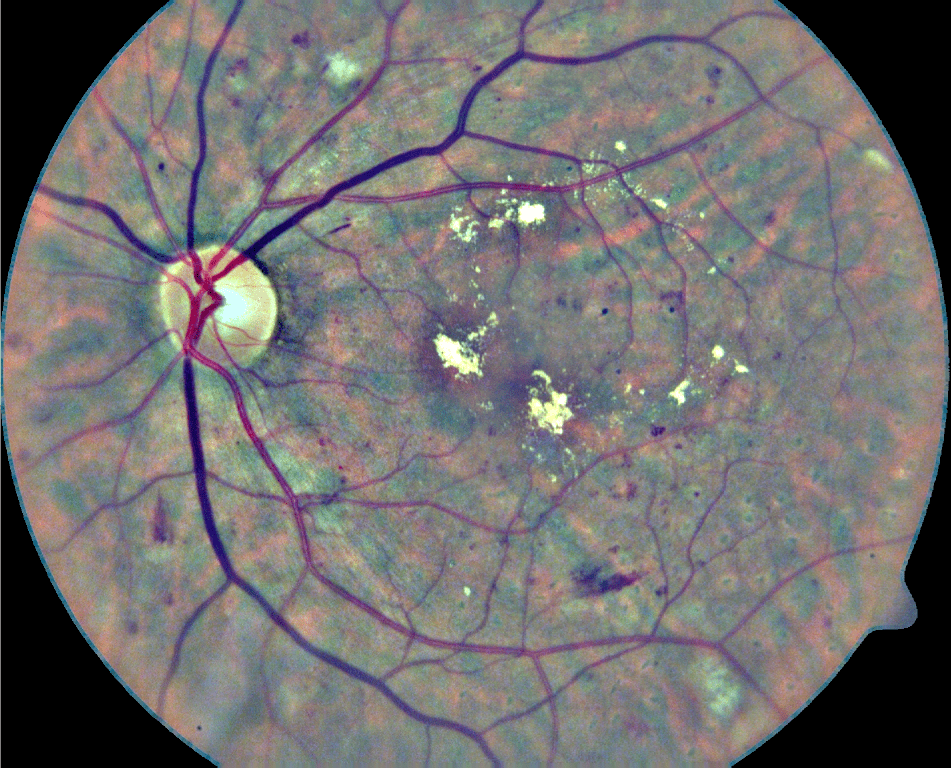}\label{fig:qualitative-image}}
  	\subfigure[Ground truth.]{\includegraphics[width=0.47\columnwidth]{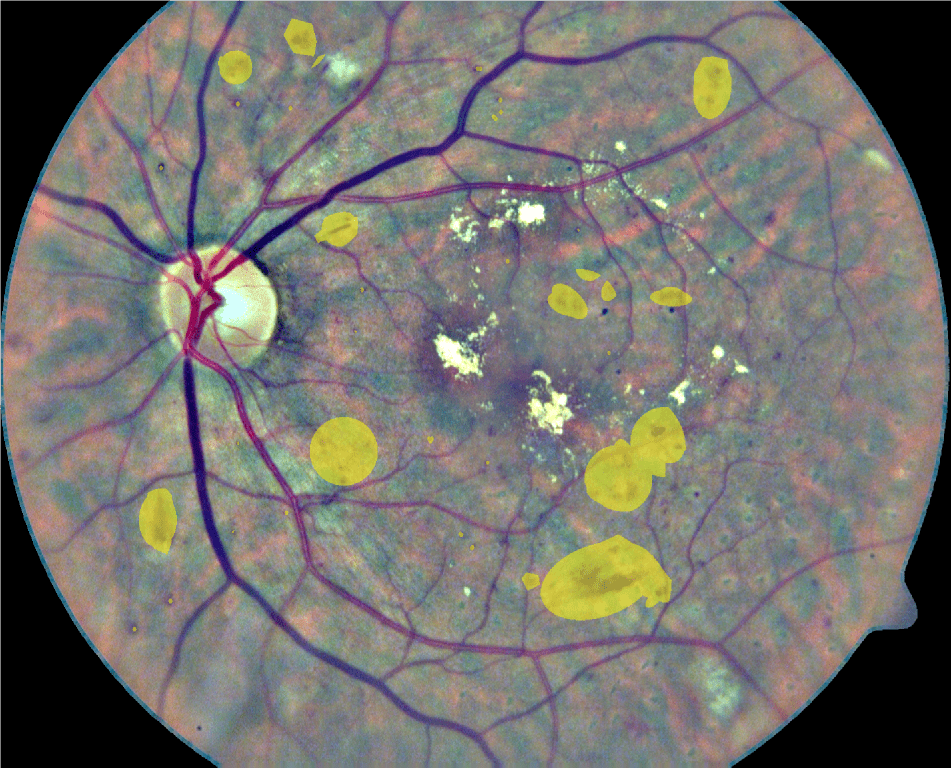}\label{fig:qualitative-ground-truth}}
    
    \subfigure[Red lesion detection]{\includegraphics[width=0.52\columnwidth]{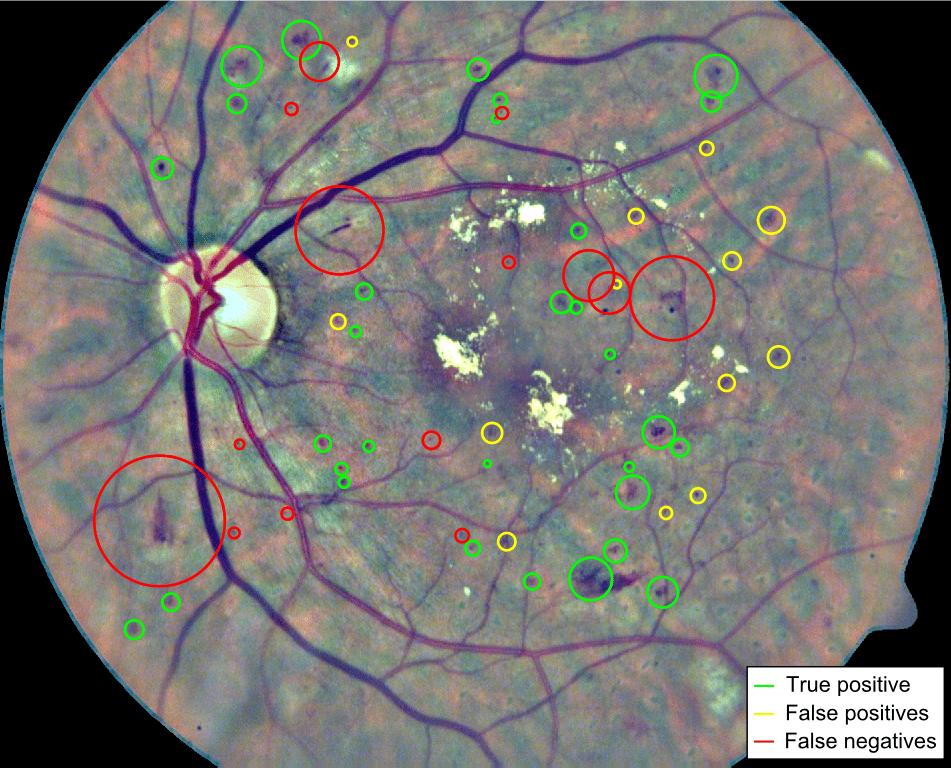}\label{fig:qualitative-detection}}
    \subfigure[Detail from (c)]{\includegraphics[width=0.405\columnwidth]{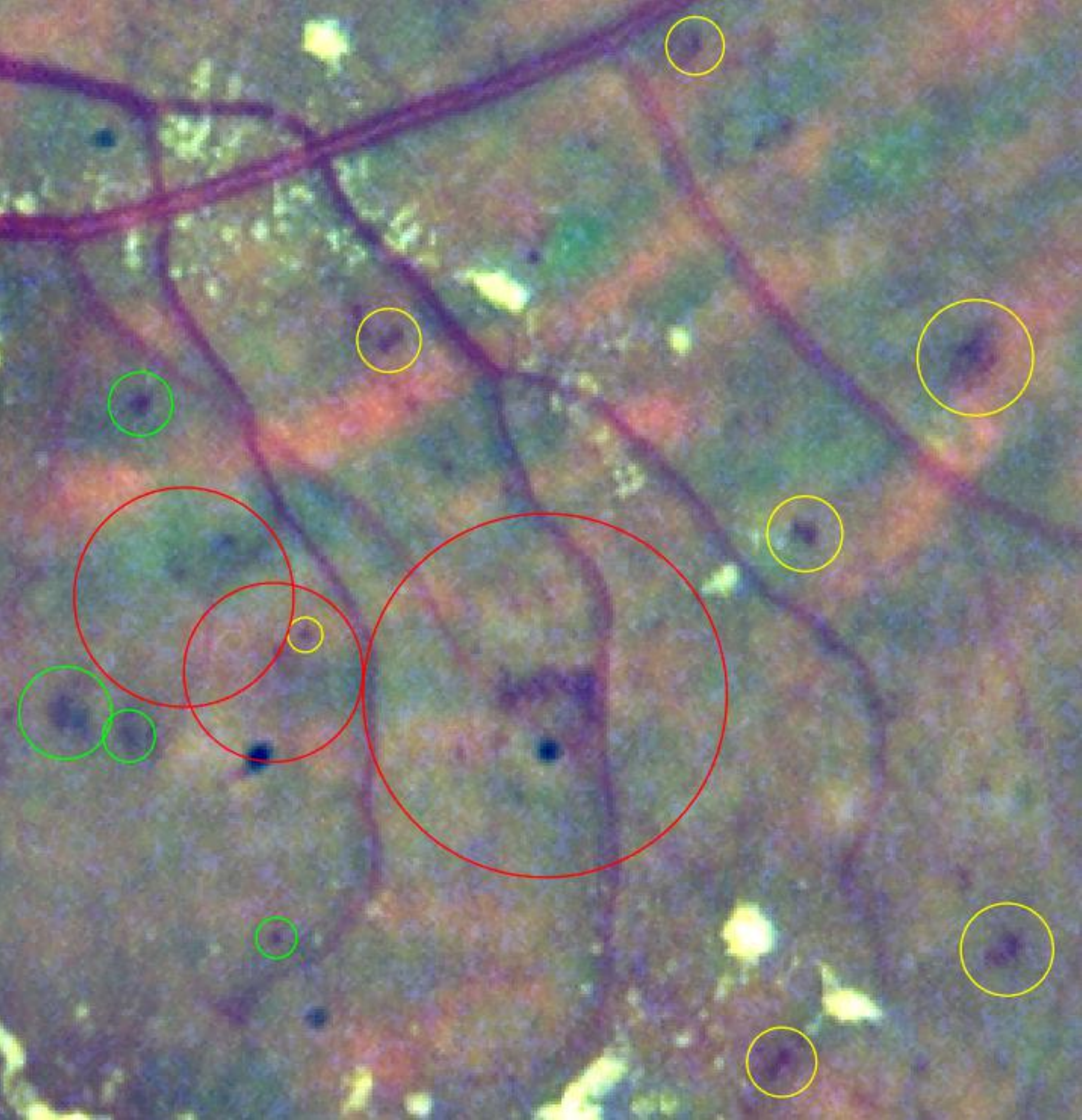}\label{fig:qualitative-detail}}

 	\caption{Qualitative results. (a) image015 from the DIARETDB1 test set. (b) Ground truth labeling at a $>25\%$ level agreement. (c) Red lesion detections obtained by thresholding the probabilities at 0.644, which corresponds to an average FPI value of 1. (d) Detail from (c) showing lesions unlabeled on the ground truth but identified by our method. }
 	\label{fig:qualitative}
 \end{figure}

Finally, qualitative results for a randomly selected image in the DIARETDB1 test set are depicted in Figure~\ref{fig:qualitative}. Green circles are detected lesions according to the ground truth labeling provided in the data set, while yellow circles correspond to lesions detected by our method but that are not labeled in the ground truth. Finally, red circles surround the lesions that were manually annotated as true lesions but were ignored by   the method. Qualitatively, many of the yellow circles appear to be microaneurysms or hemorrhages that were not detected during manual labeling due to their subtle appearance in the original RGB image.

\subsection{Per image evaluation}
\label{subsec:per-image-evaluation}

Two different experiments were conducted on MESSIDOR to estimate the performance of our method on a per image basis, one focused on detecting patients with DR, and a second based on detecting those need for immediate referral to a specialist. In both cases, we used the model learned from Experiment~1. 

\begin{figure}[t!]
  \centering
  \subfigure[Performance of DR screening]{\includegraphics[width=0.7\columnwidth]{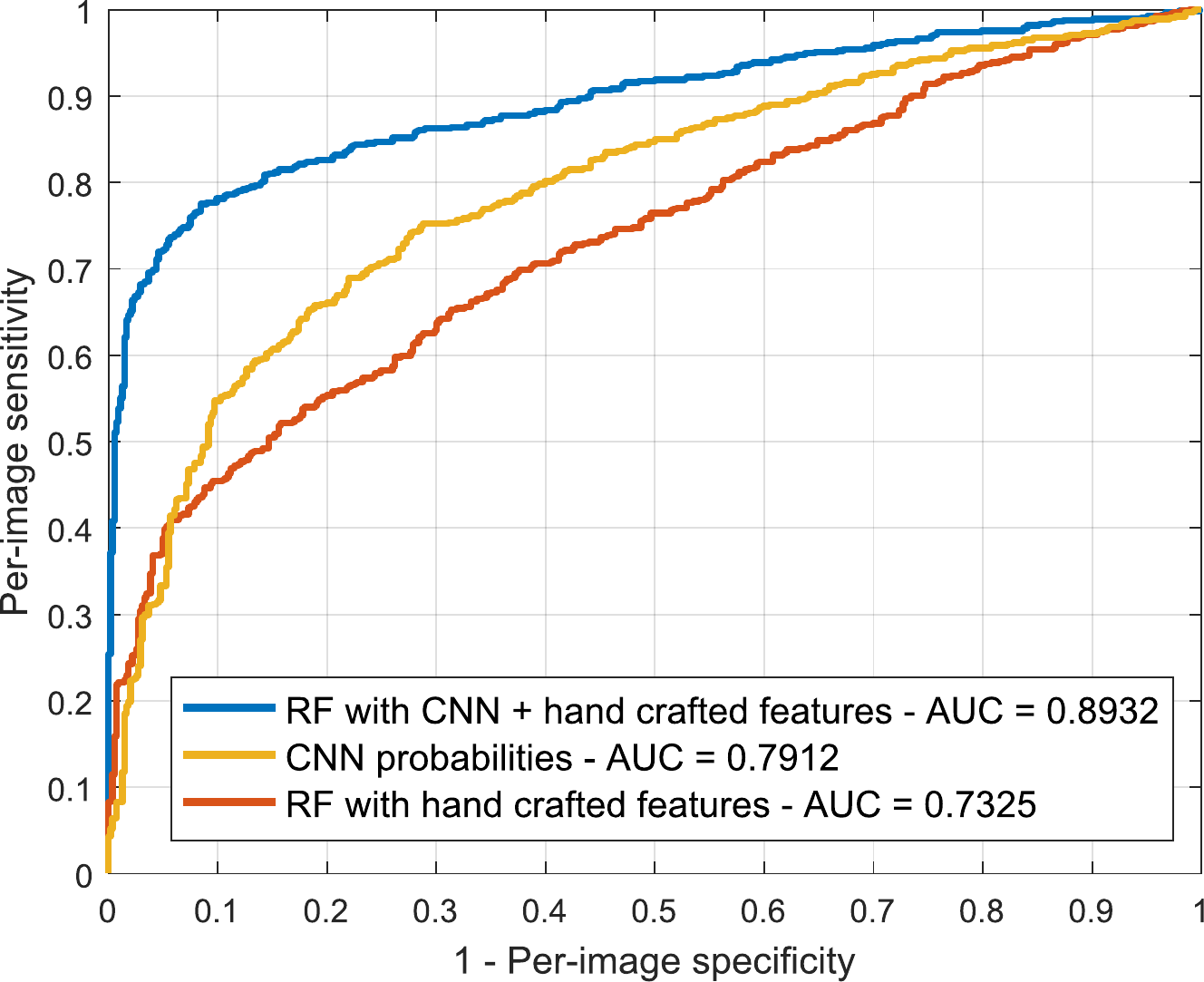}\label{fig:screening}}
  
  \subfigure[Performance of detecting patients that need referral]{\includegraphics[width=0.7\columnwidth]{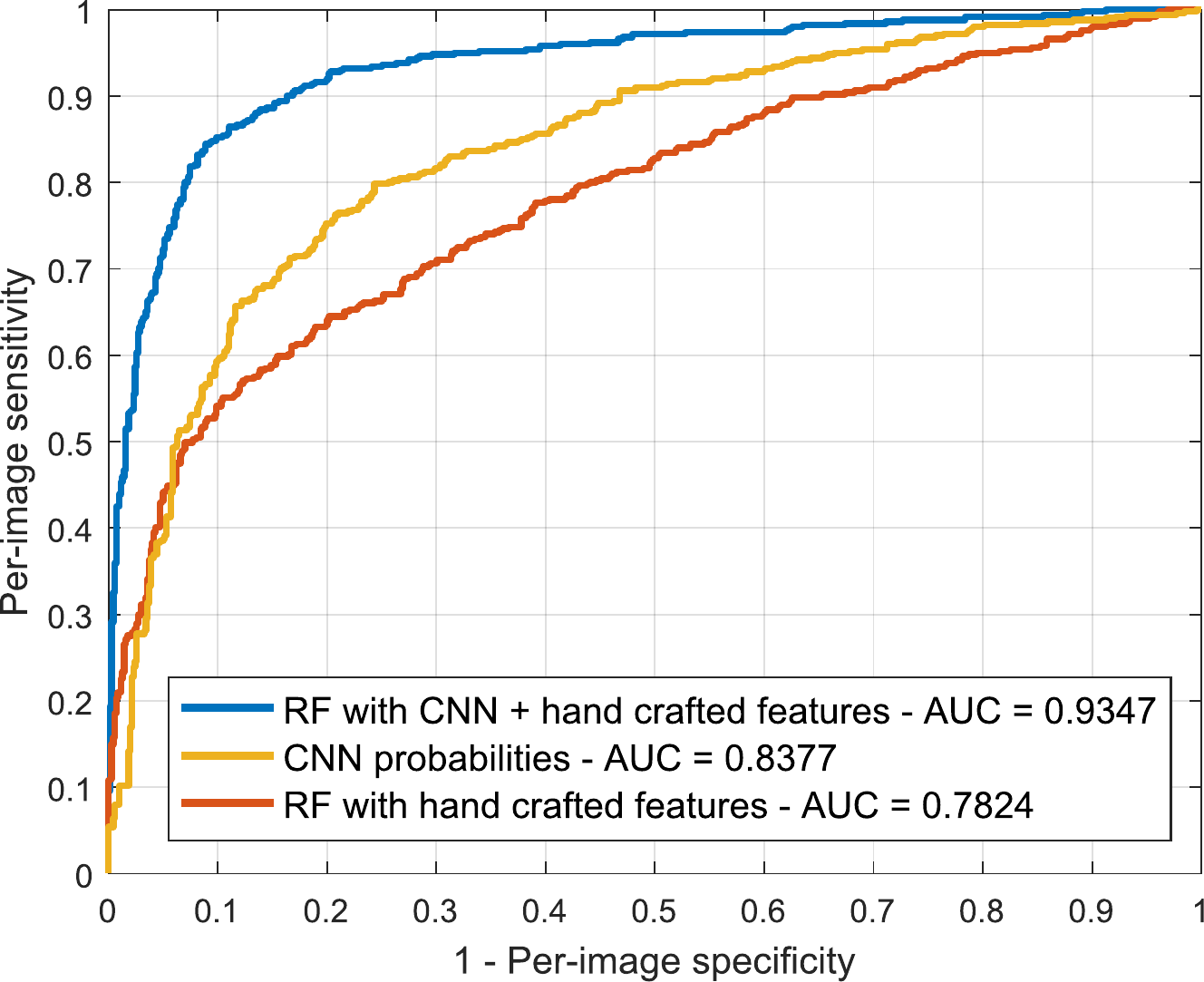}\label{fig:need-to-referral}}
  \caption{Per image evaluation. ROC curves for (a) DR screening (R1 vs. R2, R3 and R4) and (b) need for referral (R1 and R2 vs. R3 and R4) on the MESSIDOR data set.}
  \label{fig:roc-screening}
\end{figure}

Figure~\ref{fig:screening} illustrates the ROC curves for DR screening on MESSIDOR, obtained using our hybrid representation and each of the approaches separately. CNN results were obtained using the network as a classifier. A series of Mann{-}Whitney $U$ tests ($\alpha=0.05$) were performed to study the statistical significance of the differences in the AUC values~\citep{hanley1982meaning}. CNN features (AUC $= 0.7912$) perform significantly better ($p < 1 \times 10^{-3}$) than hand crafted features (AUC $= 0.7325$) for this specific task, and the combination of both sources of information results in a substantially higher AUC value of 0.8932 ($p < 1 \times 10^{-6}$). Figure~\ref{fig:need-to-referral} shows analogous behavior for detecting patients that need  referral, with the CNN performing better than the hand crafted features ($p < 2 \times 10^{-3}$), and the combined approach outperforms both individual techniques ($p < 1 \times 10^{-6}$). 

\begin{figure}[t!]
  \centering
  \includegraphics[width=0.7\columnwidth]{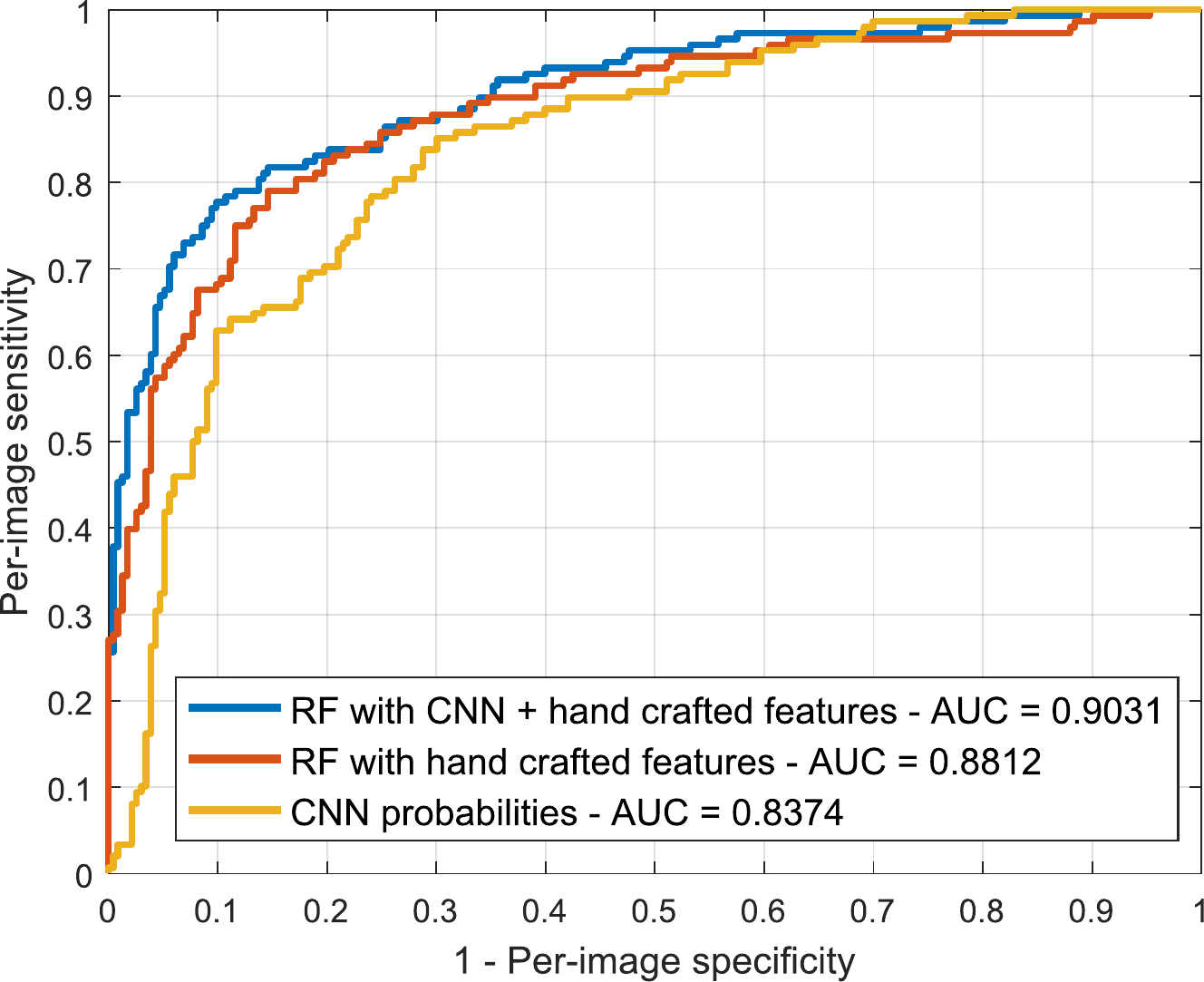}
  \caption{Per image evaluation on e-ophtha. ROC curve for DR screening.}
  \label{fig:roc-screening-e-ophtha}
\end{figure}

Our combined approach shows an analogous behavior when evaluating on e-ophtha for DR screening, as illustrated in Figure~\ref{fig:roc-screening-e-ophtha}. Our combined approach retrieved a significantly higher AUC value (0.9031) than the one reported by the CNN (AUC = 0.8374, $p < 5 \times 10^{-3}$) and the RF classifier trained with hand crafted features (AUC = 0.8812). Hand crafted features perform better than the CNN for screening in this data set, although the difference is not statistically significant according to the Mann-Whitney $U$ test.

A comparison with respect to other state of the art strategies is presented in Table~\ref{table:auc-messidor}. The performance obtained by two human experts, as reported by~\cite{sanchez2011evaluation}, is also included in the table. The results of the baseline method by~\cite{seoud2015red} were obtained using DIARETDB1 as a training set. The other methods included are either based only on red lesion detection or complemented by other features such as image quality assessment or the detection of exudates and neovascularizations.



\begin{table}[t!]
\centering
\caption{Comparison of DR screening and need of referral performance on the MESSIDOR data set. $Se$ values correspond to those obtained at a $Sp = 50\%$.}
\label{table:auc-messidor}
\resizebox{\columnwidth}{!}{
\begin{tabular}{ l | c | c || c | c }
  \hline                       
  \multirow{2}{*}{\textbf{Method}} & \multicolumn{2}{c||}{\textbf{Screening}} & \multicolumn{2}{c}{\textbf{Need for referral}} \\
  & \textbf{AUC} & \textbf{Se} & \textbf{AUC} & \textbf{Se} \\
	\hline
	\hline
	\textit{Expert A} \citep{sanchez2011evaluation} 	& 0.9220  & 0.9450 & 0.9400 & 0.9820  \\
	\hline
	\textit{Expert B} \citep{sanchez2011evaluation} 	& 0.8650  & 0.9120 & 0.9200 & 0.9760  \\
	\hline
	\hline
	\cite{antal2012ensemble} 		& 0.8750 & - & - & -  \\
	\hline
    \cite{costa2016smartphone}  & 0.8700 & - & - & - \\
    \hline
	\cite{giancardo2013validation} 	& 0.8540  & - & - & -  \\
    \hline
    \cite{nandy2016incremental} & - & - & 0.9210 & - \\
    \hline
	\cite{pires2015beyond} 	& -	& - & 0.8630 & -  \\
	\hline
	\cite{sanchez2011evaluation} 	& 0.8760	& \textbf{0.9220} & 0.9100 & 0.9440  \\
	\hline
	\cite{seoud2015red} (DIARETDB1) & 0.844	& - & - & -	 \\  
    \hline
    \cite{vo2016new} (I) & 0.8620 & - & 0.8910 & - \\
    \hline
    \cite{vo2016new} (II) & 0.8700 & - & 0.8870 & - \\
	\hline
    \hline
	\textbf{HCF} 	& 0.7325  & 0.7645 	& 0.7824 &	0.8283  \\
    \hline
	\textbf{CNN} 	& 0.7912  & 0.8471 	& 0.8377 &	0.9102  \\
     \hline
	\textbf{HCF + CNN} 	& \textbf{0.8932}  & 0.9109 & \textbf{0.9347} & \textbf{0.9721}		  \\   
	\hline  
\end{tabular}}
\end{table}

\subsection{Feature assessment}
\label{subsec:feature-assessment}

\begin{figure}[t!]
  \centering
  \subfigure[Experiment 1]{\includegraphics[width=0.45\columnwidth]{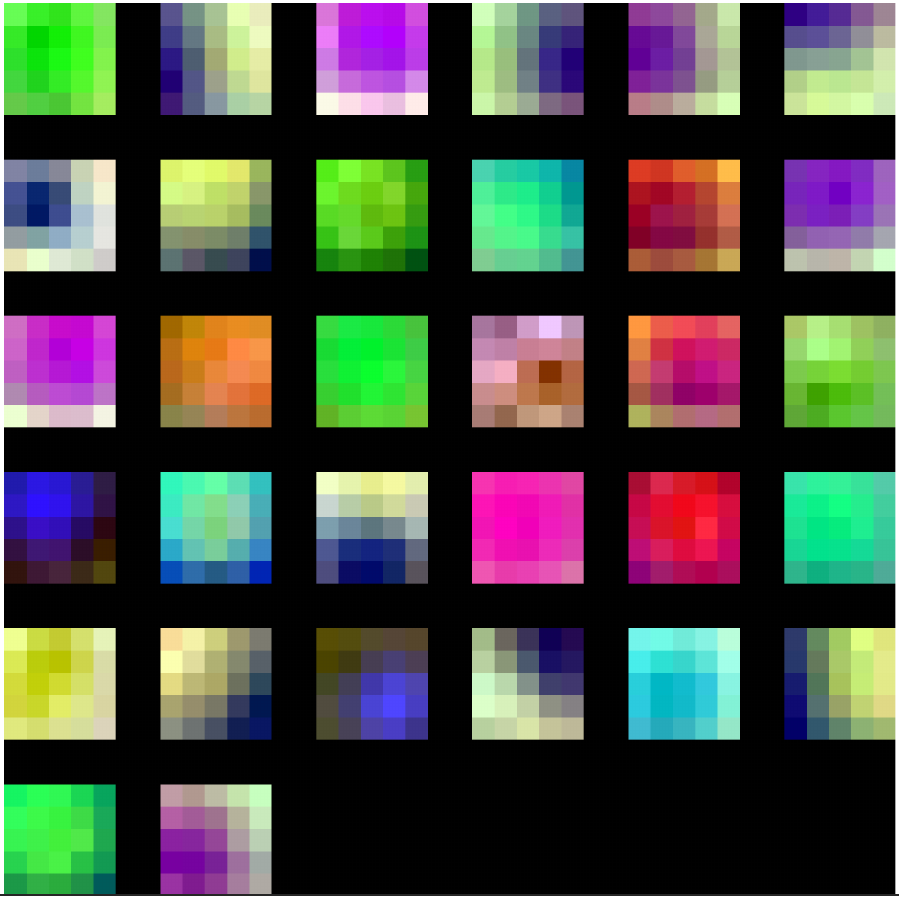}\label{fig:weights-1}}
    \subfigure[Experiment 2]{\includegraphics[width=0.45\columnwidth]{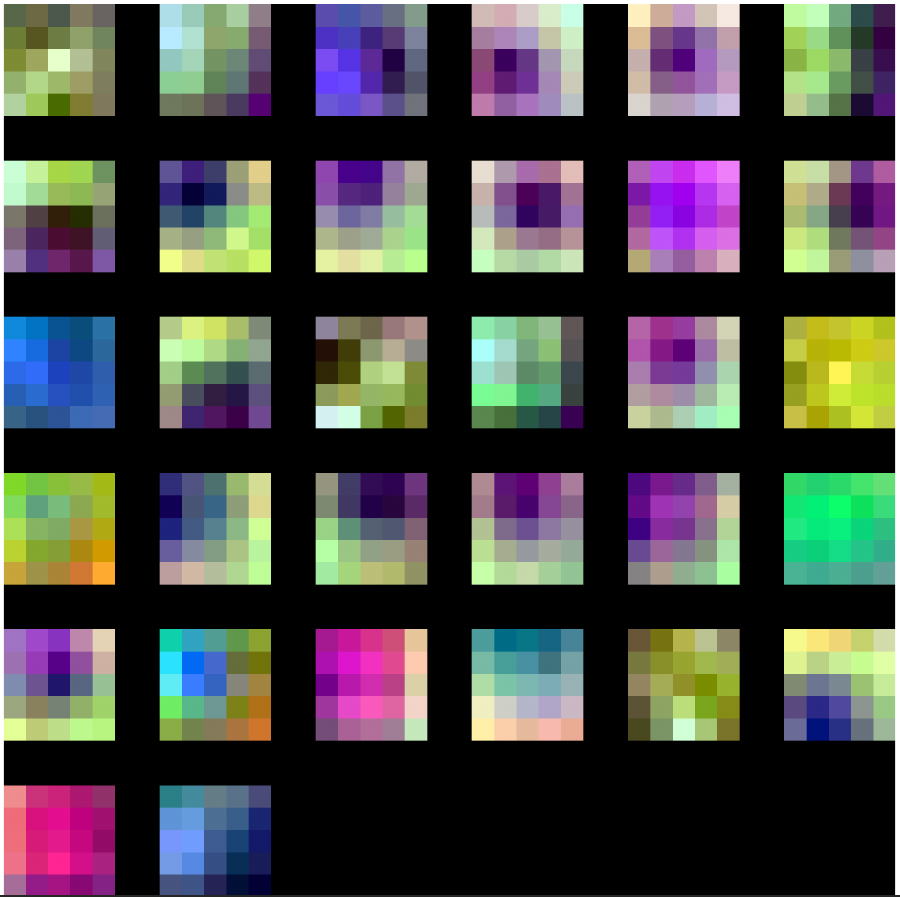}\label{fig:weights-2}}
  \caption{Learned filters on the first layer of our CNN, as obtained for each experiment in Table~\ref{table:experiments}:(a) Experiment 1 (training on DIARETDB1 training set), (b) Experiment 2 (training on DIARETDB1 and ROC training sets).}
  \label{fig:learned-filters}
\end{figure}

In order to assess the visual appearance of the deep learned features, a graphical representation of the 32 filters of size $5 \times 5 \times 3$  learned on the first layer of the CNN is presented in Figure~\ref{fig:learned-filters}. These representations allow to verify which types of high level characteristics are detected by the first layer of the network \citep{zeiler2014visualizing}. Thus, they are suitable to confirm if the network was trained for long enough, as well-trained CNNs usually display smooth filters without noisy patterns, as in this case. From Figure~\ref{fig:weights-1} it is possible to see that filters learned in Experiment 1 are mostly descriptors of the color properties of the lesions. This setting is in line with the fact that the training set used in this case contains not only small MAs but also medium size HEs, which can be more easily described in terms of their internal color homogeneity rather than their edges, which significantly varies from one to another. Other filters are able to capture purple, ellipsoidal structures corresponding to true lesions like those illustrated in Figure~\ref{fig:true-positives}. This last type of filter is more common in the first layer of the CNN learned in Experiment 2 (Figure~\ref{fig:weights-2}, which might be associated with the smaller true positive structures observed in the training set built with ROC and DIARETDB1 MAs.

\begin{figure*}[t!]
  \centering
  \subfigure[CNN features]{\includegraphics[width=0.47\textwidth]{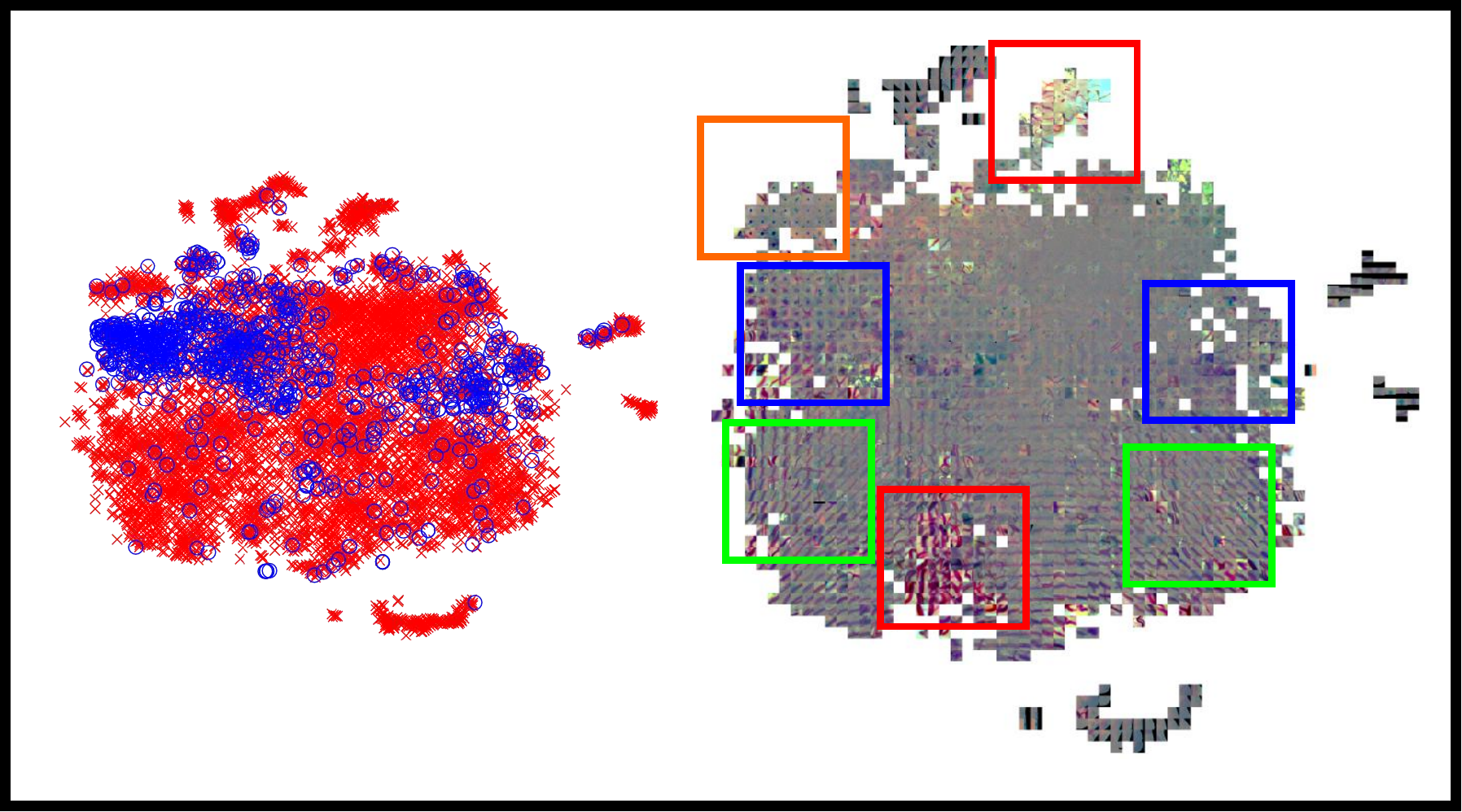} \label{fig:embedding-cnn}}
  \subfigure[Hand crafted features]{\includegraphics[width=0.47\textwidth]{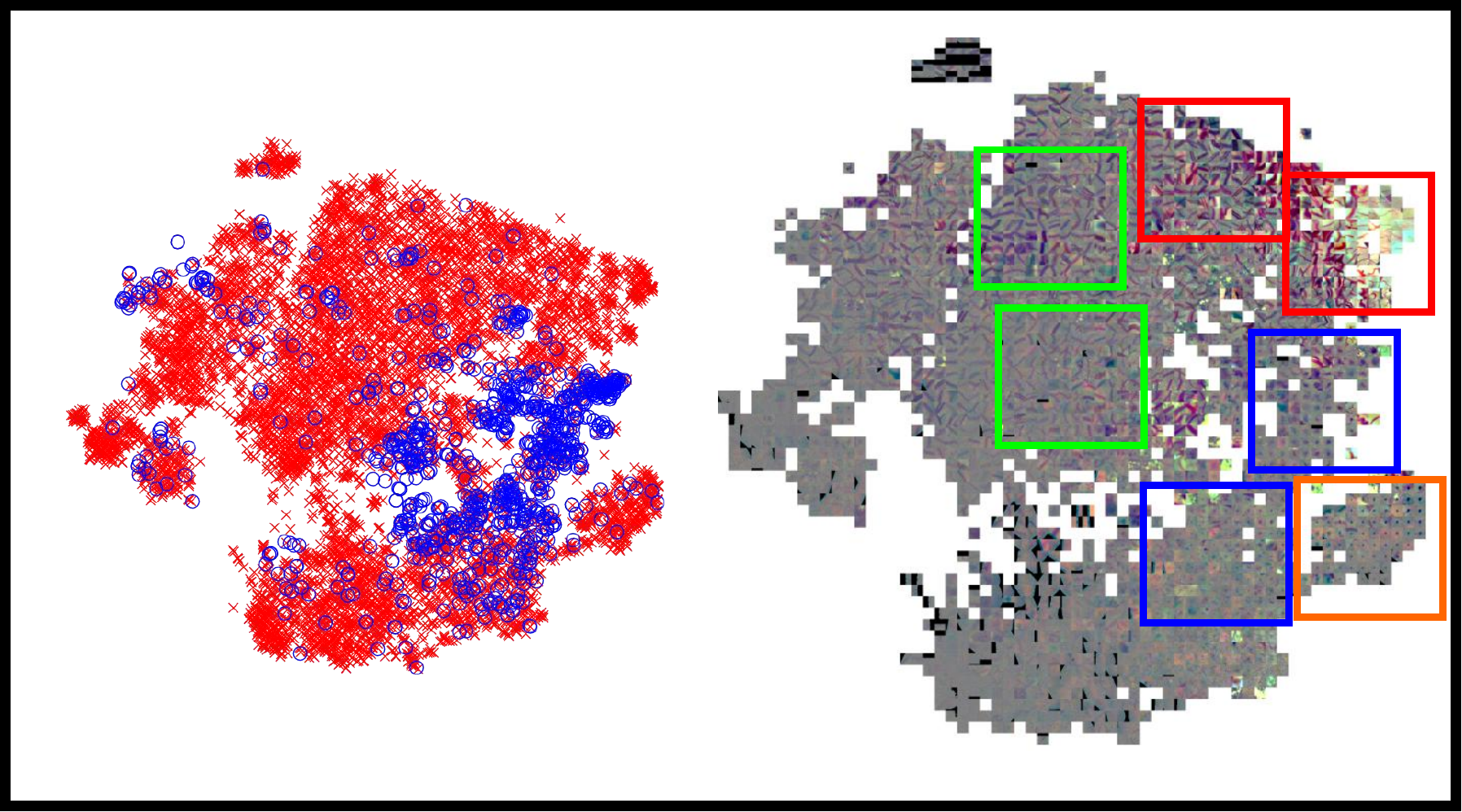} \label{fig:embedding-hcf}}
    
  \subfigure[Combined approach]{\includegraphics[width=0.47\textwidth]{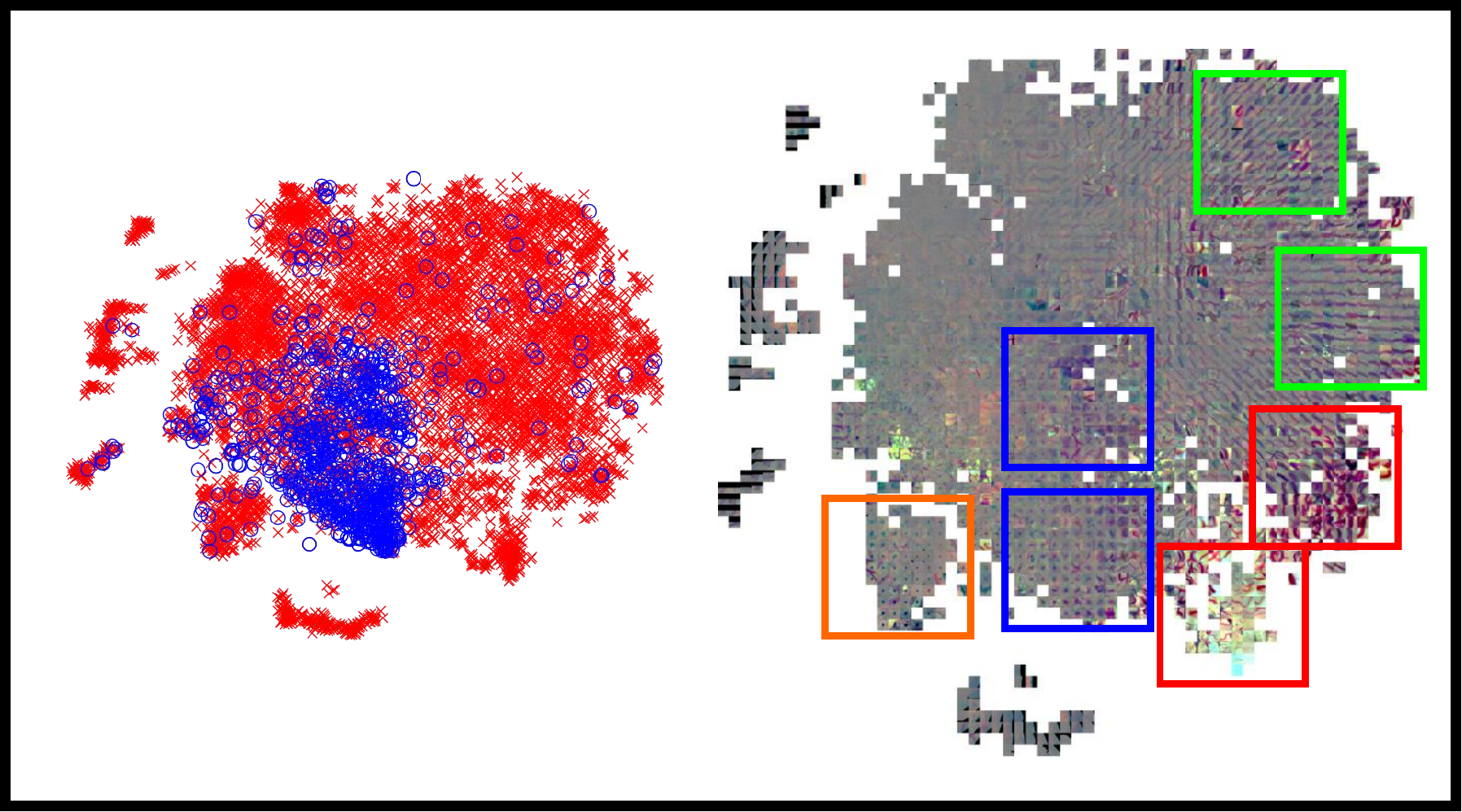} \label{fig:embedding-combined}}  
  \subfigure{\includegraphics[width=0.47\textwidth]{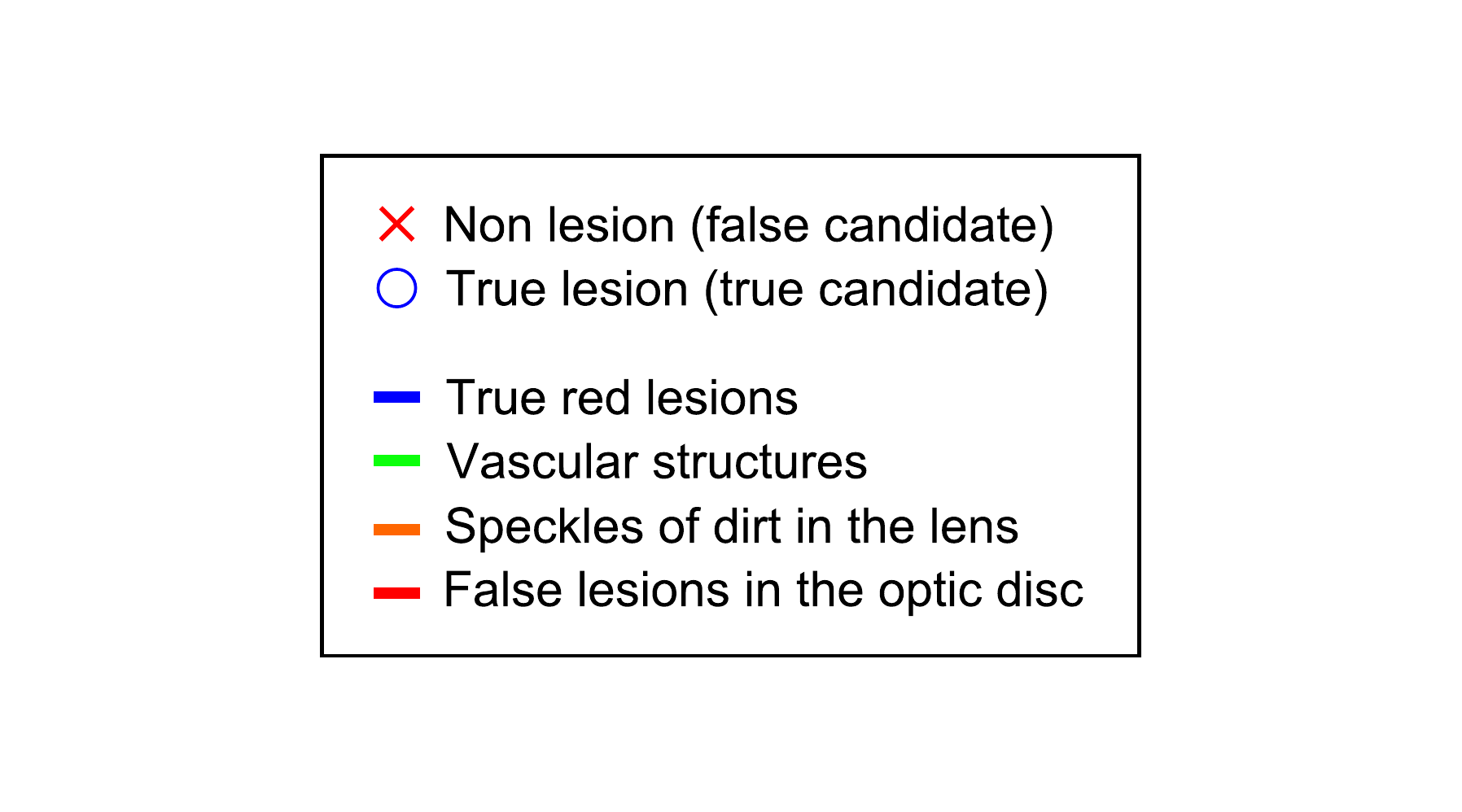} \label{fig:embedding-legend}}
  
  \caption{The $t$-SNE visualization of the patches from DIARETDB1 test set as mapped using (a) the deep learned features, (b) the hand crafted features and (c) our hybrid feature vector. Left side: color coded labels for each test sample. Right side: patches around the candidates, as visualized using the $t$-SNE mappings. Details for different types of lesion candidates are shown in Figure~\ref{fig:t-sne-detail}.}
  \label{fig:t-sne}
\end{figure*}

CNNs can be interpreted as models that transform the input images into a feature representation in which classes can be separated by the linear model in the last layer. The topology of such a space will depend on the ability of the deep learning features to characterize the inputs. Hence, if features are sufficiently good to differentiate each type of input, at least two well separated regions would be visually identified. Due to the high dimensionality of the feature space, a method is needed to embed multidimensional vectors in a 2D space, while preserving the pairwise distances of the points. The $t$-distributed stochastic neighbor embedding ($t$-SNE) was recently introduced for this purpose \citep{van2014accelerating}. We followed this approach to study the complementarity of each characterization method, and to qualitatively assess how their integration contribute to improve their original discrimination ability.  Figure~\ref{fig:t-sne} presents the $t$-SNE mappings of the DIARETDB1 test samples for each characterization approach and for our combined feature vector. The CNN descriptors corresponds to those learned in Experiment 1. The figure also includes a visual representation of the organization of the patches in the embedding space. In general, it is possible to see that the ensemble approach groups the majority of the true positive candidates within a single neighboring area. By contrast, the individual characterization strategies are not able to achieve a single cluster but rather obtain two--in the case of the deep learned features--or more--using the hand crafted features. 

\begin{figure}[t!]
  \centering
  \subfigure[CNN features]{\includegraphics[width=0.3\columnwidth]{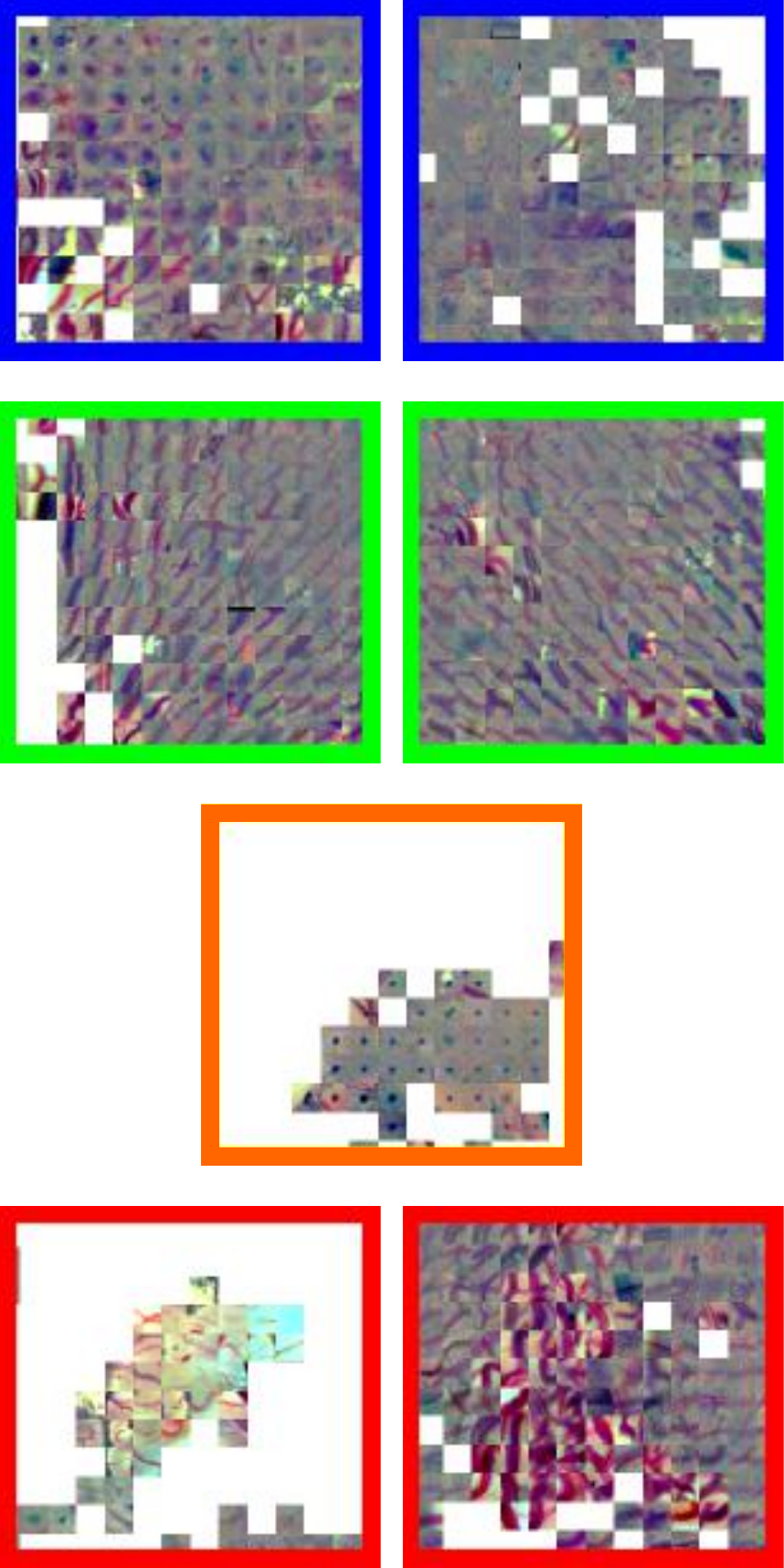} \label{fig:detailed-embedding-cnn}}
  \subfigure[Hand crafted features]{\includegraphics[width=0.3\columnwidth]{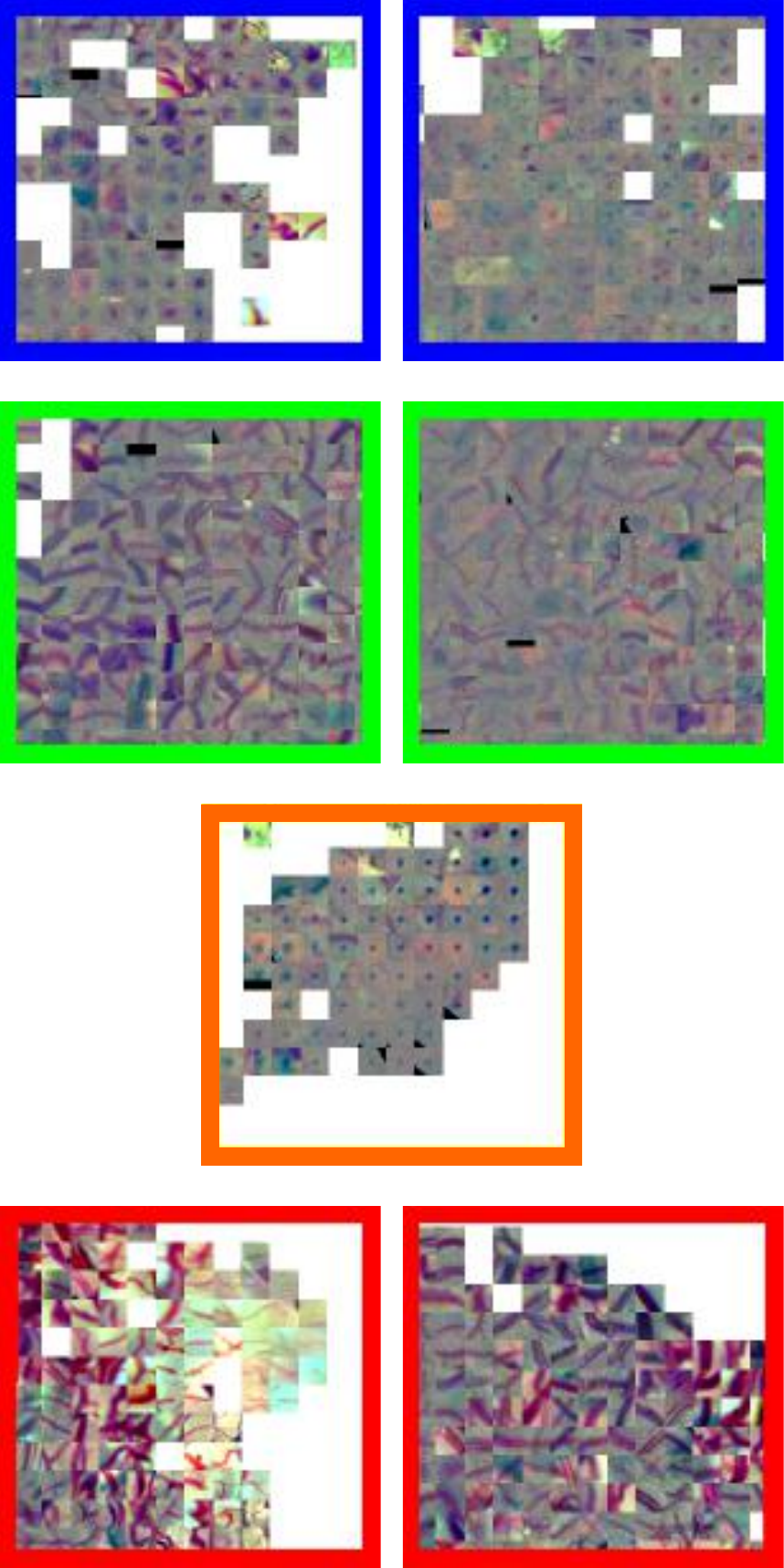} \label{fig:detailed-embedding-hcf}}
  \subfigure[Combined approach]{\includegraphics[width=0.3\columnwidth]{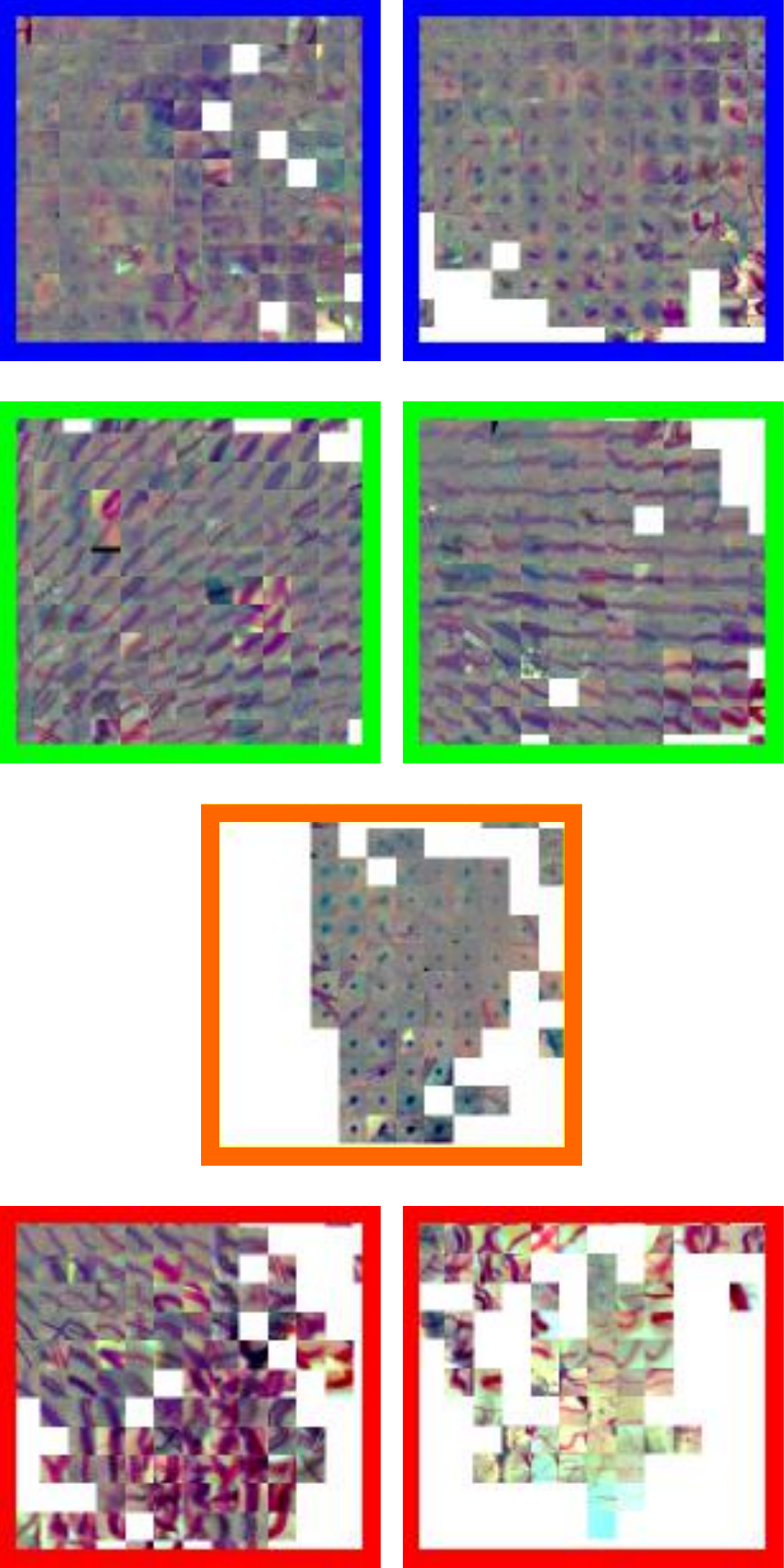} \label{fig:detailed-embedding-combined}}
  
  \subfigure{\includegraphics[width=0.4\columnwidth]{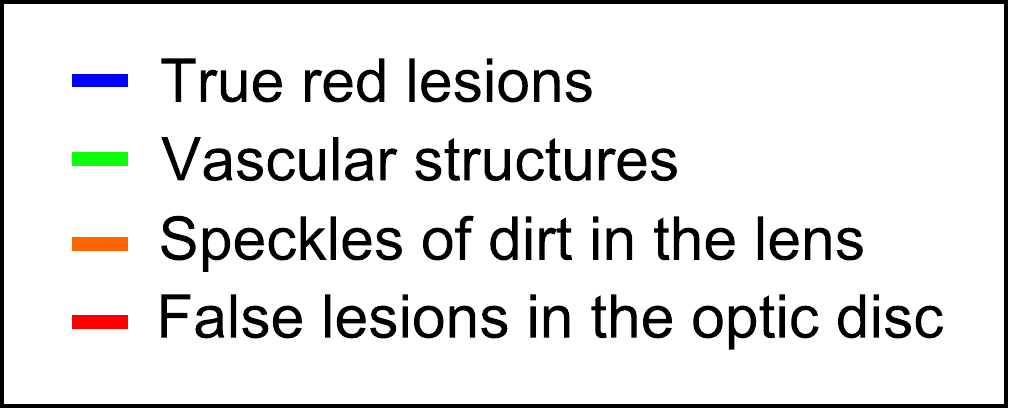} \label{fig:detailed-embedding-legend}}
  
  \caption{Details from the $t$-SNE visualization in Figure~\ref{fig:t-sne} for different types of red lesion candidates (true lesions, vascular structures, speckles of dirt in the lens and false detections in vessel curves in the optic disc): (a) Deep learned features, (b) Hand crafted features, (c) Combined approach.}
  \label{fig:t-sne-detail}
\end{figure}

Detailed regions of the embeddings are depicted in Figure~\ref{fig:t-sne-detail}. This allows better visualization of particular scenarios such as the patches around the true red lesions, the false positive candidates located in the vascular structures, the artifacts due to speckles of dirt in the lens--which are typical of the images in DIARETDB1--and the false detections within the optic disc. In general, it is possible to observe that CNN features are able to better characterize the orientation and the visual appearance of the true lesion candidates, while the hand crafted features can detect the less obvious lesions under low contrast conditions. The ability of the CNN features to discriminate orientations are more evident when dealing with vascular structures. The hand crafted approach, by contrast, is only able to capture the overall size of the vessels and their intensity properties. When combining both strategies, the main advantages of each of them are maintained. The robustness against artifacts is evident for both the deep learning based and the hand crafted features, as these false positive candidates are grouped together into separate clusters from the true lesions. A similar behavior is observed when dealing with false candidates within the optic disc area.

\section{Discussion}
\label{sec:discussion}

In general, the integration of both the deep learned and the hand crafted features  significantly improved results compared to using either approach separately. In a per lesion evaluation, the combined approach achieved a consistently higher CPM value both in the e-ophtha and DIARETDB1 test sets, and also a higher per lesion sensitivity for FPI=1, which corresponds to a clinically relevant number of false positives \citep{niemeijer2010retinopathy}. These values are also higher than those obtained by two recently published baseline methods that were evaluated on the same data set. A similar behavior is observed when evaluating the method on a per image basis. The combined approach improved the performance obtained by each characterization approach separately, meaning that the integration of both sources of information obtains a better characterization of the lesion candidates and, consequently, a more accurate detection of the individual lesions. This is supported by the extensive analysis presented in Section~\ref{subsec:feature-assessment}. Despite the fact that sufficiently deep CNNs are known to be able to learn any function of arbitrary complexity, the lack of data with lesion-level annotations does not allow our network to identify the same properties that the hand crafted features do. Nevertheless, in the analysis of the $t$-SNE mapping presented for each method (Figures~\ref{fig:t-sne} and~\ref{fig:t-sne-detail}) it is possible to see that the CNN has the ability to characterize fine-grained details such as the orientation of the lesion that are ignored by the manually selected descriptors. On the other hand, the hand crafted features have the ability to discriminate other low contrast lesions (Figure~\ref{fig:t-sne-detail}), specially hemorrhages (Figure~\ref{fig:lesion-hemorrhages}). As a result, the ensemble approach is able to outperform each individual alternative, improving performance for detecting both MA and HE simultaneously. Due to the high cost of accurately annotating small lesions, we hypothesize that this observation will continue to stand in the near future.

Results on the per image evaluation also showed that the proposed strategy is able to achieve higher AUC values than other approaches for DR screening and need-for-referral detection. 
Moreover, the methods included in Table~\ref{table:auc-messidor} are based not only on red lesion detector \citep{antal2012ensemble,giancardo2013validation,seoud2015red} but also on additional features such as the assessment of the image quality \citep{sanchez2011evaluation} and/or the presence of other pathological structures such as exudates and neovascularizations \citep{sanchez2011evaluation, pires2015beyond, costa2016smartphone}. Compared with respect to all these approaches, our method achieved a higher AUC value. Furthermore, it performed better than the DR grading method by \cite{vo2016new}, which uses fine tuned CNNs trained on a data set with 50.000 images with image level annotations. An almost equal performance was obtained for DR screening compared with the recently published method by \cite{quellec2016automatic}, which reported an AUC$=0.893$ in the MESSIDOR data set. However, such an approach uses multiple images per patient, contextual information and clinical records to learn diagnostic rules from a data set with 12.000 examinations. Our method is able to achieve a slightly higher AUC value without including any additional clinical information. Furthermore, a competitive $Se$ value was obtained in comparison with Expert B \citep{sanchez2011evaluation}, indicating that this approach can match the ability of a human observer for DR screening and detecting patients that need referral. Thus, our automated red lesion detection system could be integrated in a more general DR screening platform to improve the ability to detect DR patients. In particular, methods such as the one proposed by \cite{gulshan2016development}, which is able to identify moderate/worse and severe DR cases, can be aided by the incorporation of a red lesion detection so that the early stages of the disease can also be determined. Moreover, a reliable DR likelihood can be complemented by an indication of the abnormal areas, allowing physicians to better identify the clinical signs of the disease and to have more comprehensive feedback from the system. Furthermore, incorporating other modules for detecting other pathological structures can eventually improve the reported performance.

It is also important to underline that all the stages in the proposed method have parameters that are automatically adjusted to each image resolution. Their values, which are reported in Section~\ref{subsec:model-selection}, were empirically selected using different data than that used for evaluation, and were proportionally scaled in the subsequent experiments to compensate resolution changes. This simple approach provides an approximate scale invariance that is valuable to facilitate the adaptability of the method to be applied on images obtained using different fundus cameras.

When analyzing each individual characterization approach, it is possible to see in Experiment~1 that both the RF trained with hand crafted features and the CNN achieved higher per lesion sensitivities than the method by~\cite{seoud2015red} ($p < 2 \times 10^{-18}$ and $p < 2 \times 10^{-4}$, respectively). This is likely due to the fact that our method for extracting candidates differs from the one used by the alternative approach. Moreover, \cite{seoud2015red} eliminate the lesion candidates occurring within an estimated area around the optic disc center, which is determined using an automated approach. As a consequence, if the diameter of the optic disc is accidentally overestimated by such a method, candidates within valid regions will be suppressed and it will not be possible to recover them afterwards during the classification stage. As seen in Figures~\ref{fig:t-sne} and~\ref{fig:t-sne-detail}, our combined approach is able to discriminate the candidates within the optic disc area and the vascular structures. Hence, instead of using a rigid elimination step based on optic disc segmentation, we let the classifier to decide whether a candidate is actually a true positive or a false positive occurring on an anatomical region. This approach increases the maximum achievable per lesion sensitivity on each image, allowing to train our classifier with a larger amount of false positive lesions and to get a higher sensitivity in test time. A similar observation can be made from the results of Experiment~2, in which the hand crafted features and the deep learning based approach reported higher per lesion sensitivities than those reported by \cite{wu2017automatic}. It must be underlined, also, that the \cite{wu2017automatic} method was trained on the first half of the images with pathologies on e-ophtha and evaluated on the second half, rather than trained on a separate data set and evaluated on the complete set, as in our case. Moreover, it is worth noting that the images of the healthy patients were also included during evaluation to get a more accurate estimation of its actual performance on a real, clinical scenario.

On a per image basis, it is possible to see that the individual approaches trained in Experiment~1 are not able to achieve AUC values higher than those reported by \cite{seoud2015red} (Table \ref{table:auc-messidor}). This is likely due to the fact that, as indicated by the authors, their method is more accurate for detecting blot HEs and MAs than HEs with other shapes. The images in MESSIDOR were originally graded as R0 and R1 taking into account the number of MAs (Table~\ref{table:messidor}) \citep{decenciere2014feedback}. Hence, being more accurate in the detection of MAs will result in a better ability to distinguish much earlier stages. When individually using the hand crafted features or the CNN, both methods are less precise for detecting MAs but better for discriminating other HEs. This argument is supported by results presented in Figure~\ref{fig:per-lesion-type}, in which it is possible to see that the per lesion sensitivity values obtained for MA detection are lower than those reported for HEs. Moreover, it was observed that the CNN performed equally or better than the RF trained with manually engineered features on the low FPI regime for MA detection. This explains the behavior observed in Figure~\ref{fig:roc-screening}, where the CNN probabilities achieved a higher AUC value for DR screening and need for referral detection. Nevertheless, the combination of both approaches with the RF classifier consistently improved their individual performance, achieving a much better characterization of the MAs (as observed in the improvements reported in Figure~\ref{fig:lesion-microaneurysms}) and, consequently, a better discrimination of the DR patients.


\section{Conclusions}
\label{sec:conclusions}

We have proposed a novel method for red lesion detection in fundus images based on a hybrid vector of both CNN-based and hand crafted features. A CNN is trained using patches around lesion candidates to learn features automatically, and those descriptors are complemented using domain knowledge to improve their discrimination ability. Results on benchmark data sets empirically demonstrated that the resulting system achieves a new state-of-the-art in this domain, and that combining both sources of information provides statistically significant improvements compared to using each of them separately. A similar behavior is observed when evaluating our screening system both for DR and need-for-referral detection, reporting higher AUC values than those obtained by other existing approaches based not only on red lesion detection but also on analyzing other pathologies such as bright lesions or neovascularizations, or even learning classifiers using additional clinical information. Considering the high cost of manually labeling fundus photographs at a lesion level, our method represents a robust alternative to improve performance of other deep learning based approaches. An open source implementation and the detection masks are made publicly available at \sourcecodeurl. \\ 

\noindent \textbf{Acknowledgements}\\

This work is supported by Internal Funds KU Leuven, FP7-MC-CIG 334380, an Nvidia Hardware Grant and ANPCyT PICT 2014-1730, PICT Start-up 2015-0006 and PICT 2016-0116. J.I.O. is funded by a doctoral scholarship granted by CONICET. We would also like to thank Dr.\ Seoud for her assistance. \\

\noindent \textbf{Conflicts of interest}\\

The authors declare that there are no conflicts of interest in this work.

\section*{References}
\bibliographystyle{plainnat}
\bibliography{RetinalImagingReport.bib}

\end{document}